\documentclass[conference]{IEEEtran}
\IEEEoverridecommandlockouts
\usepackage{cite}
\usepackage{amsmath,amssymb,amsfonts}
\usepackage{algorithmic}
\usepackage{graphicx}
\usepackage{textcomp}
\usepackage{xcolor}
\def\BibTeX{{\rm B\kern-.05em{\sc i\kern-.025em b}\kern-.08em
    T\kern-.1667em\lower.7ex\hbox{E}\kern-.125emX}}
\usepackage[colorlinks=true, urlcolor=black, linkcolor=black, citecolor=black]{hyperref}
\usepackage{soul,color}
\usepackage{multirow}
\usepackage{varwidth}
\usepackage{ctable}

\newcommand{\commentsEnabled}{1}            
\newcommand{\commentSai}{}
\newcommand{\commentPablo}{}
\ifnum \commentsEnabled=0
   \renewcommand{\commentSai}[1]{}
   \renewcommand{\commentPablo}[1]{}
\fi
\ifnum \commentsEnabled=1
    \renewcommand{\commentSai}[1]{\textcolor{green}{#1}}
    \renewcommand{\commentPablo}[1]{\textcolor{blue}{#1}}
\fi

\graphicspath{{Figures/}}
\begin{document}

\makeatletter
\newcommand{\linebreakand}{%
  \end{@IEEEauthorhalign}
  \hfill\mbox{}\par
  \mbox{}\hfill\begin{@IEEEauthorhalign}
}
\makeatother

\title{An Open and Reconfigurable User Interface to Manage Complex ROS-based Robotic Systems
\thanks{The research leading to these results has received funding from the European Union’s Horizon 2020 research and innovation program under grant agreement n° 870133, correspondent to the project entitled REMODEL, Robotic Technologies for the Manipulation of Complex Deformable Linear objects.}
}

\author{\IEEEauthorblockN{Pablo Malvido Fresnillo}
\IEEEauthorblockA{\textit{FAST-Lab, Faculty of Engineering}\\
\textit{and Natural Sciences,} \\
\textit{Tampere University} \\
Tampere, Finland \\
pablo.malvidofresnillo@tuni.fi}

\and
\IEEEauthorblockN{Saigopal Vasudevan}
\IEEEauthorblockA{\textit{FAST-Lab, Faculty of Engineering}\\
\textit{and Natural Sciences,} \\
\textit{Tampere University} \\
Tampere, Finland \\
saigopal.vasudevan@tuni.fi}

\and

\linebreakand

\IEEEauthorblockN{Jose A. Perez Garcia}
\IEEEauthorblockA{\textit{Design and Fabrication in Industrial} \\
\textit{Engineering, University of Vigo} \\
Vigo, Spain \\
japerez@uvigo.es}

\and


\IEEEauthorblockN{Jose L. Martinez Lastra}
\IEEEauthorblockA{\textit{FAST-Lab, Faculty of Engineering}\\
\textit{and Natural Sciences,} \\
\textit{Tampere University} \\
Tampere, Finland \\
jose.martinezlastra@tuni.fi}
\and

}

\maketitle

\begin{abstract}
The Robot Operating System (ROS) has significantly gained popularity among robotic engineers and researchers over the past five years, primarily due to its powerful infrastructure for node communication, which enables developers to build modular and large robotic applications. However, ROS presents a steep learning curve and lacks the intuitive usability of vendor-specific robotic Graphical User Interfaces (GUIs). Moreover, its modular and distributed nature complicates the control and monitoring of extensive systems, even for advanced users. To address these challenges, this paper proposes a highly adaptable and reconfigurable web-based GUI for intuitively controlling, monitoring, and configuring complex ROS-based robotic systems. The GUI leverages ROSBridge and roslibjs to ensure seamless communication with ROS systems via topics and services. Designed as a versatile platform, the GUI allows for the selective incorporation of modular features to accommodate diverse robotic systems and applications. An initial set of commonly used features in robotic applications is presented. To demonstrate its reconfigurability, the GUI was customized and tested for four industrial use cases, receiving positive feedback. The project's repository has been made publicly available to support the robotics community and lower the entry barrier for ROS in industrial applications.
\end{abstract}

\begin{IEEEkeywords}
ROS, Human-robot interaction, Graphical user interface, Web-based programming, Industrial automation
\end{IEEEkeywords}
\section{Introduction}
\label{sec:intro}

A robotic system encompasses all the software and hardware components that allow the automation of a process using a robot. From a hardware point of view, this can include multiple robots, end effectors, sensors, transportation systems, and other machinery. To effectively control, communicate, and coordinate all these devices, a multi-module software is typically necessary, especially when the application is complex and the robot trajectories cannot be predefined \cite{Wong2022Jun, Nam2020Jun}. In such cases, where the system requires intelligence to perceive its environment and plan movements accordingly, its software may include a plethora of modules for perception, Artificial Intelligence (AI) processing, motion planning, task planning, control, and safety management, among others. Communicating and integrating this significant amount of nodes and devices (which may belong to different vendors) would be extremely complicated using the robot proprietary software. Aiming to simplify this task, a generic robot framework called Robot Operating System (ROS) was presented in 2007 \cite{MalvidoFresnillo2023Oct}. 

ROS is an open-source middleware suite that provides a set of tools (i.e., topics, services, and actions) to facilitate the communication between the multiple modules of a system \cite{FresnilloICPS}; attributes that make it an ideal framework to develop complex and extensive robotic applications. As a consequence, numerous software libraries and packages have been developed in ROS to help build robotic applications, such as MoveIt, for motion planning; RVIZ, to visualize the 3D robot environment; Gazebo, a physics engine to simulate the robot operation; ROS Industrial, for vendor-agnostic robot communication; or FlexBE, a user-friendly engine to program complex high-level robot behaviors. However, despite its considerable advantages, ROS remains primarily utilized in research. The main reason for this is its steep learning curve, making it very difficult for individuals without robotics expertise to learn and use, which is the case for many operators and end users \cite{Ajaykumar2021Oct}. Furthermore, the management of complex robotic systems, composed of big networks of interconnected nodes, using the terminal can become very challenging even for experienced users.

The integration of Graphical User Interfaces (GUIs) presents a promising avenue to improve the accessibility and usability of ROS, allowing novel and advanced users to interact with complex systems in an easy and intuitive way \cite{Oulasvirta2020Feb}. In fact, several GUIs tailored for specific ROS libraries and applications have been developed, significantly enhancing the user experience, as indicated by surveys and users' feedback \cite{Coleman2014Apr}. Examples of this are the RVIZ plugin of MoveIt, used for motion planning, or the FlexBE User Interface, used for defining high-level robot behaviors. However, no complete and generic high-level user interfaces for controlling, monitoring, and configuring entire complex ROS-based robotic systems have been presented.
Hence, the aim of this paper is to develop such an interface, which will allow seamless interaction with ROS, making it more accessible for all kinds of users and reducing its entry barrier in industrial applications. The purpose of this is to help the robotic community, so the project's repository has been made publicly available\footnote{ \href{https://github.com/pablomalvido/UI_REMODEL}{https://github.com/pablomalvido/UI\_REMODEL}}.

The following sections of the paper are structured as follows: Section~\ref{sec:sota} analyzes the evolution of User Interfaces (UI) and reviews existing solutions to interface robotics and ROS-based systems.
Sections~\ref{sec:methodology} and \ref{sec:features} describe the proposed UI. The former presents its general architecture, while the latter focuses on the comprehensive list of modular features that were developed, which can be selectively incorporated to fit the requirements of the target application.
The adaptability and easy reconfiguration of the UI are then tested in Section~\ref{sec:UC} by implementing it in four industrial use cases.
Finally, Section~\ref{sec:conclusions} concludes the paper and introduces the intended future activities.

\section{State of the art}
\label{sec:sota}

\subsection{Evolution of User Interfaces}

Since the inception of the first computer, developers have given significant consideration towards user interfaces (UI), and Human-Computer Interactions (HCI) deals with the study of optimizing the interaction between the human and the computer system \cite{Carroll1998}. 
The first interactive (direct manipulation) interface, where a user could input information and observe the computed outputs, was developed in 1963 using a rudimentary light pen \cite{Sutherland1998Jul}. 
Consequent research led the development of the first GUI- Xerox Parc \cite{Johnson1989Oct}, at Xerox Palo Alto in 1981 \cite{Lampson1988Jan}. 
Although it was not commercially successful, this pioneering effort inspired significant computing milestones, including the creation of Apple Lisa in 1982 and the Macintosh in 1984 \cite{Myers1998Mar}.

The evolution of UIs in the coming decades capitalized on the graphical capabilities of computers, efficiently visualizing complex information for user interaction \cite{Jansen1998Apr}, and adeptly concealing back-end programming language details \cite{Martinez2011Mar}.
Moreover, GUIs have become ubiquitous across a wide range of devices today including (but not limited to) professional workstations, personal computers, smart phones, home appliances, industrial machinery and more. 
All of the aforementioned devices are built on hardware and operating systems (OS) which are proprietary, and software applications built on these platforms have low-level dependencies which are specific to the manufacturer's platform (device type and/or OS) \cite{Seffah2004Oct}.
Furthermore, the software services and their respective UIs could either possess vertical usability (interaction compatibility within a single platform) or horizontal usability (interaction compatibility across multiple platforms) \cite{Majrashi2015ACU}.
However, promoting horizontal usability by creating platform agnostic software was challenging as the developer needs to rewrite several sections of code to interface with the varying (platform-specific) libraries or toolkits, provided by every manufacturer.

Compatibility issues between software applications and various platforms can be mitigated by utilizing remote computing functionalities hosted on dedicated servers, and visualizing the output on the local devices' GUIs.
In particular, these output results can be visualized using specialized software applications that offer cross-platform compatibility.
Specifically, these software applications are web browsers, and their prevalence and capabilities have increased over the past three decades \cite{Grosskurth2006}. 
This was made possible due to the emergence of the internet and World Wide Web (WWW) along with the advances in consumer device technologies. 

Web browsers are built using HyperText Markup Language (HTML) \cite{Berners-Lee1995Nov}, which enables them to structure page layouts, insert hyperlinks, integrate multimedia, collect user inputs, and is the global standard for cross-platform compatibility.  
Along with Cascading Style Sheets (CSS) \cite{Bos2006} and JavaScript, the website's UI information could be styled and made interactive, respectively. 
This enables devices to connect to remote servers and access services from across the world, which could be universally accessed by standard web browsers, and subsequently visualized using the GUI of the device. 

\subsection{UIs for Robotics}

Robotic manipulators are becoming increasingly pervasive across multiple essential industries, which encompass (but are not limited to) food processing \cite{Iqbal2017May}, medical applications \cite{Kasina2017Oct}, automotive industry \cite{Bartos2021Jan}, horticulture \cite{Cubero2020Jul}, assisted living for the elderly \cite{Mohebbi2020Sep}, and many more \cite{Ben-Ari2017Oct}. 
As robotics are becoming more commonplace in everyday life, the prospect of non-experts having to interact with them on a daily basis is increasing. 
This situation poses a lot of uncertainty, regarding the attitude and the usage efficiency, depending on the nature of the robots status- whether it is treated as a piece of equipment or an autonomous entity which can work alongside you \cite{Latikka2021Nov}.
In the case of the former, where robots are being teleoperated or are semi-autonomous, an application specific UI is required, catering to the expertise and skill level of the target user group. 
For user groups whom have only a low to medium level of expertise with such robotic systems, the absence user-centered design can result in poor usability, long task times, high error counts, large support costs, long training times, and user dissatisfaction \cite{Henneman1999Jan}.

The UI applications could be built on local integrated development environments (IDE) such as ROBOGUIDE, Kuka Sim, MotoSim, RobotStudio, and more \cite{mikhalevich2017developing}. 
These are primarily examples of proprietary IDEs for the various industrial robot manufacturers, which require paid licenses and suitably powerful local hardware to perform optimally.
Alternately, there are several open source IDEs and frameworks for creating robotic applications, which are not limited by the licensing or the robot manufacturer, these include RoboDK, V-REP (IDEs) and ROS (framework) \cite{quigley2009ros}. 
This capability of ROS to be independent of the manufacturer specific licensing and architecture, could enable it to operate as the de facto standard platform across majority of the robot models available commercially, with the capability to build your own custom robotic manipulators.

\subsection{UIs for ROS-based systems}

While frameworks like ROS maybe open source and have an architecture adaptable to any robot configurations, the framework has several shortcomings regarding the usability, portability and platform dependency.
ROS operates on a Linux based operating system, and the default accessibility to the system is through the terminal environment \cite{Rajapaksha}. 
This creates unintended barriers to widespread use of the system, primarily because of the general populations lack of exposure to work with these environments.
In order to overcome this limitation, several user interfaces were developed to promote usability and improve the user's experience. 

A software application called TurtleUI for visualizing the TurtleBot2 manipulator was developed using python and QT, and is run locally on the host's workstation \cite{Barros}.
The primary purpose of the interface was to reduce the number of terminal tabs required to perform the various operations/tasks.
It provides basic functionalities to initialize the system, move the manipulator, stream the camera, configure settings, and run the application. 
Similarly, a software application was developed for an autonomous surface vehicle, utlizing the Qt framework \cite{Velamala}. 
The primary purpose for this UI was enable easy initialization/killing of nodes, congregate sensor data for visualizing and system feedback, executing relevant commands and manual control of the craft.
Furthermore, it enables the easy addition of new actuators and sensors to the vehicle, and control other types of autonomous vehicles with minimal modifications to the code.

ROS4HRI (ROS for Human Robot Interactions) is a locally installable framework, which aims at promoting interoperability and re-usability of ROS systems, focusing on interfacing skeleton tracking, face recognition and natural language processing (NLP) with robotic applications \cite{Mohamed2021}.
This framework builds on existing ROS functionalities, to create a human kinematic model to formulate interaction scenarios with robot manipulators. 
Additionally, it tracks four critical (not mutually exclusive) human identifiers, which are the face identifier, body identifier, voice identifier and the person identifier - which captures relevant data and creates a permanent identifier (ID) of the person.
Further research on NLP based robot interactions was conducted in project IntelBot, to handle uncertain and unknown words contained in natural language, communicate with service robots \cite{Bandara}.
A web based GUI is also developed here, to visualize the target of interest and to manually control the robot.
Similar research on NLP was implemented in drone autopilot technology, to help non-skilled pilots to have a smoother flying experience \cite{rajapaksha2019}. 
ROS is used as the middleware between the NLP subsystem and Gazebo simulator, where the flight navigation is planned.

RoboServ is a minimalist web-based UI with simple button layouts to teleoperate a mobile robot, enabling video feedback in the form of a continuously refreshing image \cite{Costa2016}.
This is a simple implementation using Python and Flask servers, and is integrated with ROS using TCP socket, and the UI is visualized using HTML and JavaScript.
Alternately, web UIs could be far more complex and filled with interactive features.
VIKI is an open-source and web-based user interface, that reduces the complexity of configuring the manipulator behavior by collecting existing ROS packages into modules, that are logically similar \cite{Hoogervorst2017May}. 
It is a software package that provides an abstraction layer between the user and the ROS system, and provides a drag and drop interface for building the system interactions.
It is built on python and traditional web development tools.
Robotic Programming Network (RPN) utilizes Robot Web Tools to write and execute ROS code for compatible robot manipulators, entirely online and execute it in real time at the robot server \cite{Casan}. 
It could visualize the feedback on the web-browser, utilizing the actionlib package of ROS.
It is successfully integrated as an extension to online learning environments, targeting education.

As expressed in the aforementioned subsections, user interfaces have evolved over the decades. 
The trend tends towards intuitive usability, interoperability, ease of access, and freedom of utilization. 
For the robotics sector, web-based ROS UIs perfectly fit this description.
The development of powerful tools enables the creation of interactive UIs, which can be accessed regardless of the location and the user's device.
The availability of open-source packages and support provided by the manufacturers and the ROS community is enabling the development of highly customizable solutions.
Our paper aims to present such a solution, which is easy to use and customize to fit the requirements of any ROS-based robotic system.
\section{UI Design and structure}\label{sec:methodology}

\setlength{\textfloatsep}{8pt}
\begin{figure*}[!t]
    \centering
    \includegraphics[width=1.6\columnwidth]{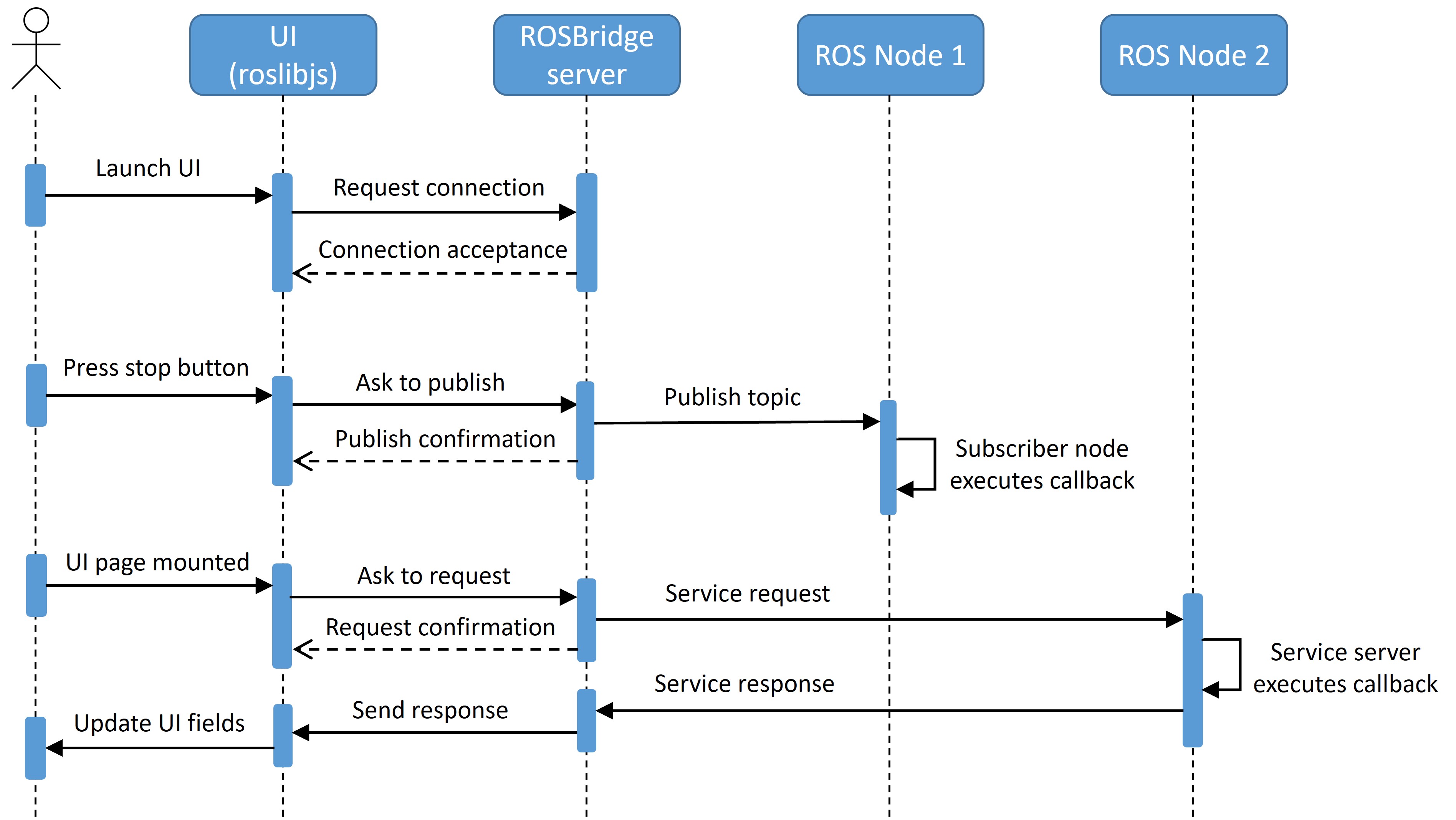}
    \caption{UML sequence diagram depicting various communication examples between the UI and a ROS system using ROSBridge and roslibjs. To emphasize the communication between the UI and ROS, the communication between the ROS nodes and the ROS Master is not represented in the diagram.}
    \label{fig:seq}
\end{figure*}

The objective of this research is to develop an universal and reconfigurable UI for ROS systems, which is generic enough to be used by any system, but at the same type can be easily adapted to attend to the specific needs of its target application. In order to accomplish this, the proposed solution is composed of a scalable platform, which is common for its implementation in any system, and a list of features, which can be integrated into the generic platform based on the requirements of each use case. Thus, the current section introduces the common UI platform, while the subsequent (Section \ref{sec:features}) documents all the developed features.

The UI has been designed as a web-based application, which allows users to access and interact with the system via standard web browsers. This approach offers flexibility, as users are not restricted by particular software installations or device limitations, allowing them to seamlessly utilize the UI across various devices and operating systems. To achieve this, the UI front-end has been developed using HTML, CSS, and JavaScript, while its back-end functionality is provided by the ROS nodes of system. To communicate them (i.e., the front-end and the back-end),  ROSBridge \cite{Crick2016Aug} and roslibjs \cite{roslibjs} are employed. ROSBridge is used to run a WebSocket server on the ROS side, which can be interfaced by an external client to establish a bidirectional communication with ROS. On the other hand, the roslibjs library provides JavaScript with a set of functions and APIs that can be used to easily interact with ROS. This allows JavaScript to create a client and manage its communication with the ROSBridge server, thus controlling the connection between the UI and ROS. Additionally, roslibjs provides JavaScript with a plethora of possibilities, such as subscribing to ROS topics, publishing messages, calling ROS services, interfacing the ROS parameter server, or listening to TF transforms, among others. Fig.~\ref{fig:seq} depicts an example of the communication between the UI and ROS. 

Regarding the framework used to construct the UI, Vue.js (Vue) was selected because of its simplicity, flexibility, and performance. Vue can be employed in two distinct manners: to develop and manage entire websites, referred to as Single-Page Applications (SPAs), or to generate standalone widgets that govern specific elements of the webpage. In this case, it was decided to build the UI as an SPA for two main reasons, its superior speed and its modularity. Unlike conventional multi-page applications, where almost every interaction involves a server request, SPAs operate within a single web page, updating content dynamically without necessitating full page reloads. This reduces significantly the loading times, as the browser only has to send one initial request to the server to get the HTML page, and then Vue can modify its content directly. Regarding its modularity, SPAs follow a component-based development approach, which facilitates code reusability, as components can be developed independently and utilized in different parts of the application. The UI leverages this methodology to favor its flexibility and reconfigurability. Hence, multiple components with distinct functionalities are developed, and then, these are selectively included in the UI based on the requirements of the system being controlled.

\setlength{\textfloatsep}{8pt}
\begin{figure*}[!t]
    \centering
    \includegraphics[width=1.9\columnwidth]{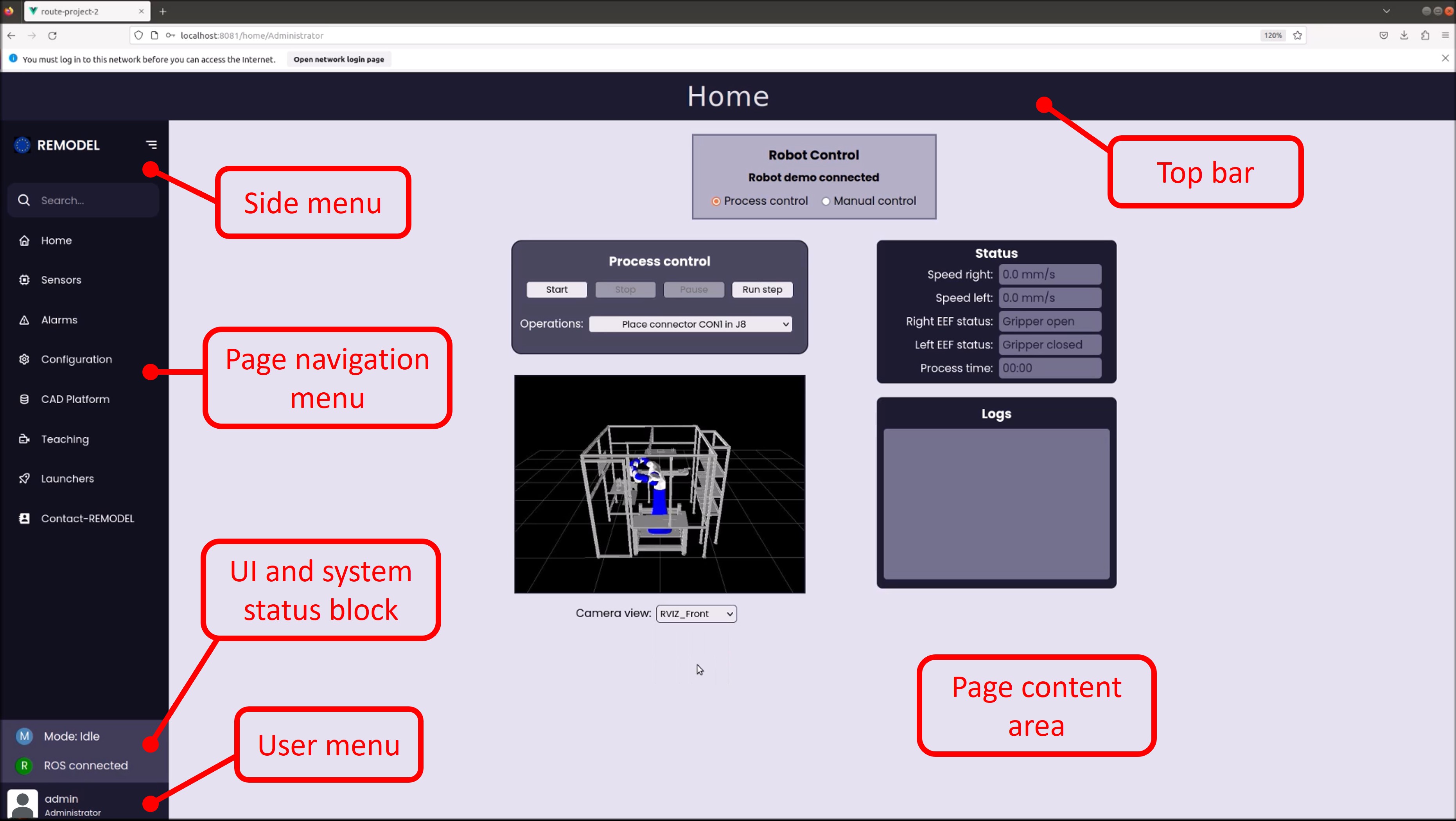}
    \caption{UI structure and common components.}
    \label{fig:structure}
\end{figure*}

To ensure the reconfigurability and scalability of the UI, a generic structure was adopted, which can easily be tailored for any use case implementation. As can be seen in Fig.~\ref{fig:structure} , this structure is composed of three components, the side menu, the top bar, and the page content area.
The side menu is an expandable Vue component composed of three elements, which has been adapted from \cite{akahon2024Apr}. The top element is a navigation menu, which allows to switch the page displayed in the UI. As mentioned earlier, the UI is an SPA, which means that it is composed of a single HTML page. Therefore, instead of requesting a new HTML page to the server, this menu injects a different view into the content area component of the current one. To avoid confusion, from now on, the manuscript will refer to the browser page retrieved from the server as \emph{HTML page}, and to the page displayed in the content area as \emph{page}.
The second element of the side menu is the UI and system status monitoring block, which provides information about the status of the socket connection with ROS, and the current operational mode of the robotic system (i.e., idle, running, alarm, programming, etc.). To keep this information current, the UI employs events to detect changes in the socket connection and a ROS subscriber to receive updates on the current operational mode.
Finally, at the bottom of the side menu, there is a user menu, which displays information about the current user, such as its name and role, and contains a logout button.

The second component of the UI is the top bar, which consists of a Vue component positioned on the top of the page that indicates the name of the displayed page. Finally, the third component is the page content area, which is the section of the UI where the current page contents are displayed. This component is therefore defined as a Vue router, which switches the displayed view and updates the UI URL when changing the page. As a result, all the pages of the UI must be developed as Vue view components.
\section{UI Features}
\label{sec:features}

As introduced in Section \ref{sec:methodology}, to facilitate the reconfigurability and adaptability of the UI, it has been built as a generic platform into which different features can be incorporated. This is possible thanks to the modularity of Vue, which allows to develop each feature as an independent component that can be easily added to the UI. Thus, besides this common platform, a comprehensive list of features has been developed, which covers some of the main necessities of robotic systems (e.g., initialization of system modules, manual and automatic robot control, sensor data visualization, etc.). All these features are presented in the subsequent subsections. It's worth noting that the UI only represents the front-end component of the system, and in most cases, additional back-end developments are needed to achieve the desired functionality. Therefore, aside from the UI aspects, these subsections also delve into some of the required ROS developments, which are finally compiled and summarized in Section~\ref{subsec:ROS}. 
Additionally, a UI demonstrator video\footnote{ \href{https://youtu.be/ZKxnzJHjA1o}{https://youtu.be/ZKxnzJHjA1o}} is provided to showcase the functionality of all these features and help the readers understand better the capabilities of the UI.

\subsection{Multirole security clearance}

This feature grants users distinct permissions and visibility within the UI based on their assigned role, which currently can be either administrator or operator. Each user is assigned a unique role, ensuring appropriate access control and segregation of duties. To access the UI, users must authenticate through the login menu (visible in Fig.~\ref{fig:login}) by providing their username and password. This login information is securely encrypted and stored in an immutable file, ensuring the confidentiality and integrity of user credentials. Thus, to add a new user to the system or modify the information of an existing one (for instance, its role), the immutability attribute of this file needs to be temporarily suspended, which requires root privileges for authorization.

\setlength{\textfloatsep}{8pt}
\begin{figure}[!t]
    \centering
    \includegraphics[width=.85\columnwidth]{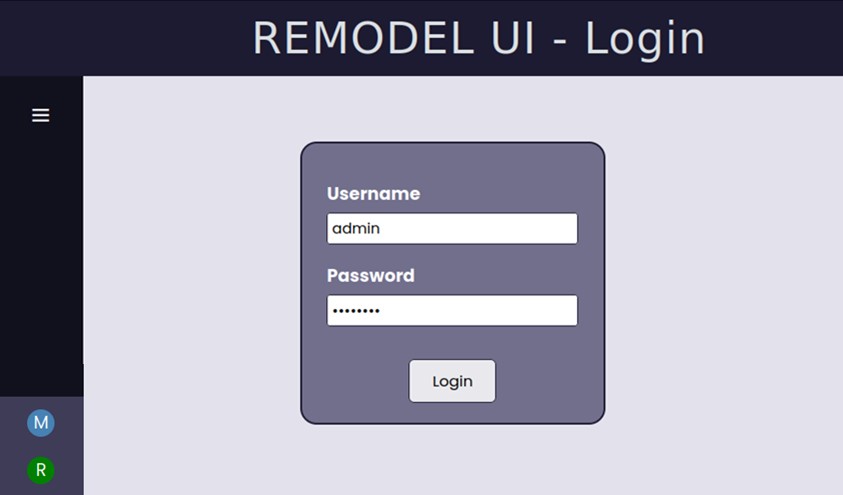}
    \caption{UI login menu.}
    \label{fig:login}
\end{figure}

After logging in, the role of the user is consulted and stored in a Vue data property. The value of this property is then used to determine which pages are displayed in the side menu, and therefore, which pages can be accessed by each role. Furthermore, the visibility of some elements within the different pages can be dependent on the value of this property. This can be used, for instance, to alter the modules displayed on the ``Launchers" page, limiting this way the system nodes that each user can initiate.
In the current implementation, only two roles are differentiated, the operator, who can only control and monitor the robot execution, and the administrator, who additionally can configure the system, define new robot trajectories and tasks, operate the robot manually, and visualize the sensor readings for maintenance or debugging purposes. However, this approach is expandable, and additional roles can be easily defined.

\subsection{Dynamic modules launching and monitoring}

As mentioned in Section~\ref{sec:intro}, robotic systems are usually composed of a significant number of modules that provide distinct functionalities. However, in most cases, not all the functionalities are needed all the time. Hence, working with all the modules constantly active may consume unnecessary resources and slowdown the system. To prevent this, this feature allows the user to selectively launch and stop the different modules of the system from the UI. Furthermore, it contains an ``All" option situated at the bottom of the page, which allows to initiate or terminate all modules simultaneously. Aside from the launch and stop buttons, each module has a status indicator, which provides feedback on its state with a color code (as can be seen in Fig.~\ref{fig:launch}). The state can be \emph{active} (green color), when all the nodes of the module are active; \emph{inactive} (gray color), when all of them are inactive; \emph{transitioning} (orange color), when the module has been launched but not all its nodes are active yet, or when it has been stopped and not all its nodes are inactive yet; or \emph{incomplete} (red color), when only part of the module nodes are active, either because they have died or because they haven't been launched. The logic for the state transitions is depicted in Fig.~\ref{fig:states}. To increase the robustness, this policy accounts for situations where the user might be using the terminal at the same time to initiate or terminate nodes.

\setlength{\textfloatsep}{8pt}
\begin{figure}[!t]
    \centering
    \includegraphics[width=.99\columnwidth]{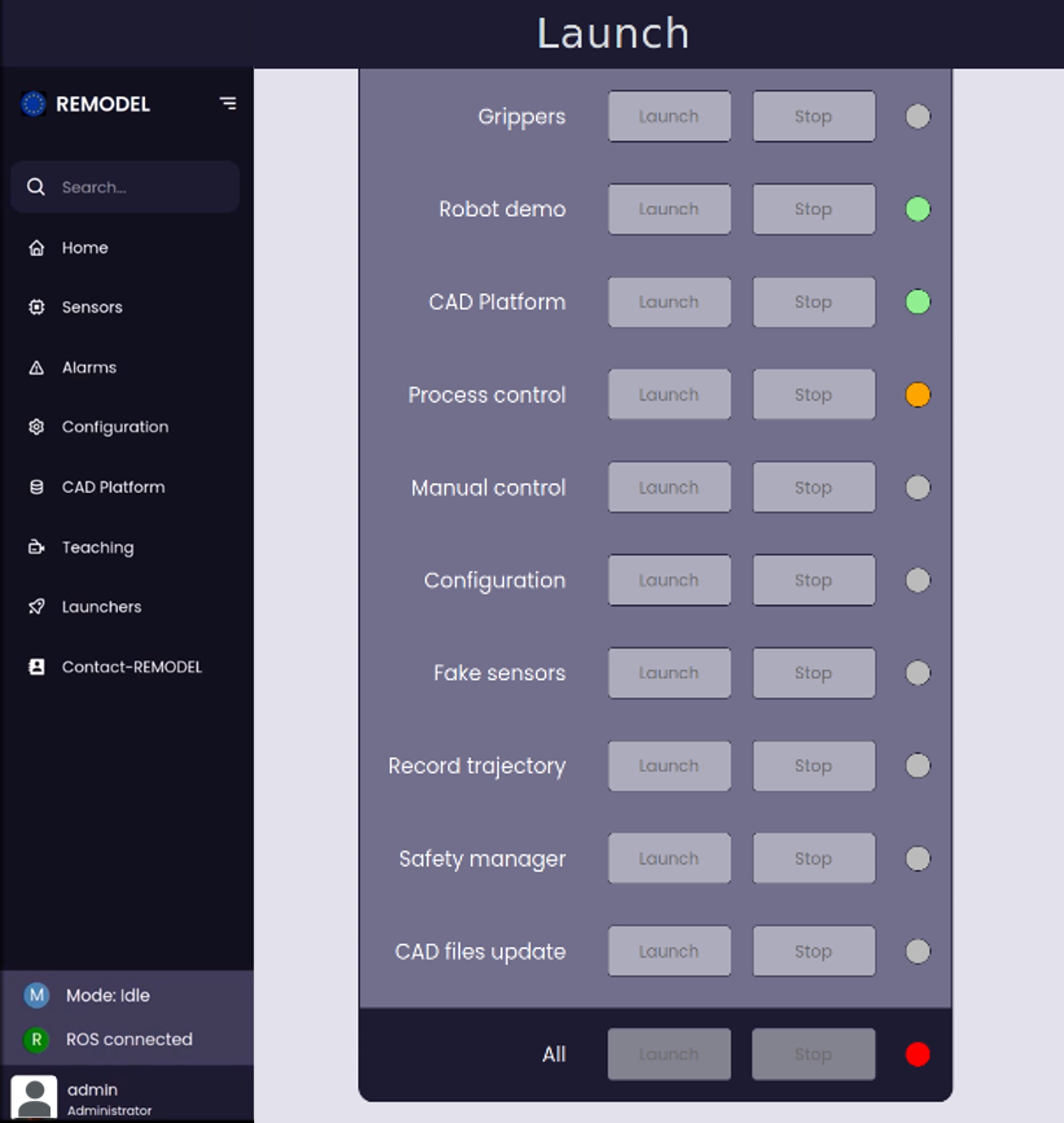}
    \caption{UI “Launchers” page. All the Launch and Stop buttons are disabled because the “Process control” module is being launched.}
    \label{fig:launch}
\end{figure}

\setlength{\textfloatsep}{8pt}
\begin{figure}[!t]
    \centering
    \includegraphics[width=\columnwidth]{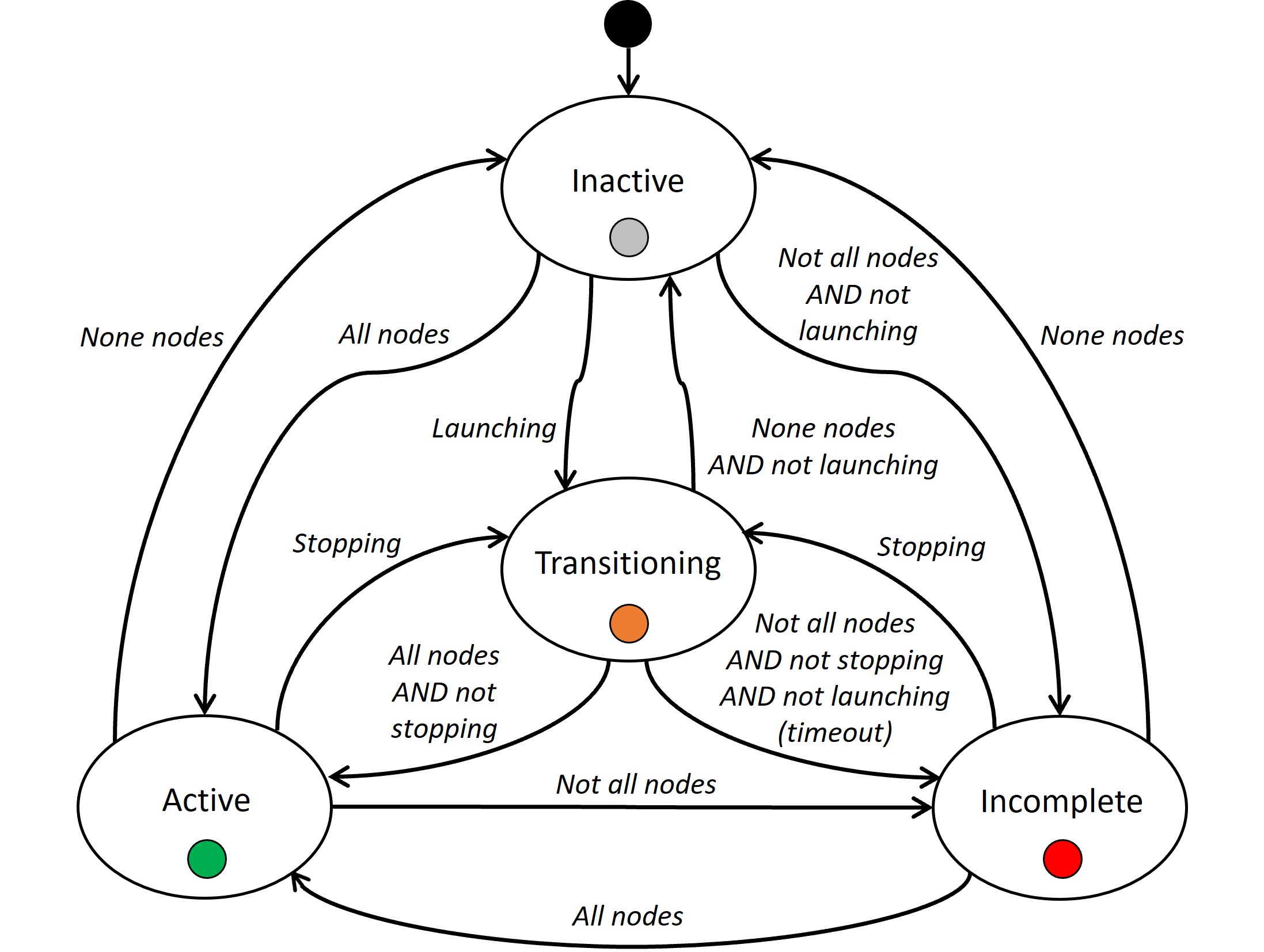}
    \caption{UML state machine diagram of the module state transitions. The list of active ROS nodes is checked every second.}
    \label{fig:states}
\end{figure}

\setlength{\textfloatsep}{8pt}
\begin{figure*}[!t]
    \centering
    \includegraphics[width=.9\textwidth]{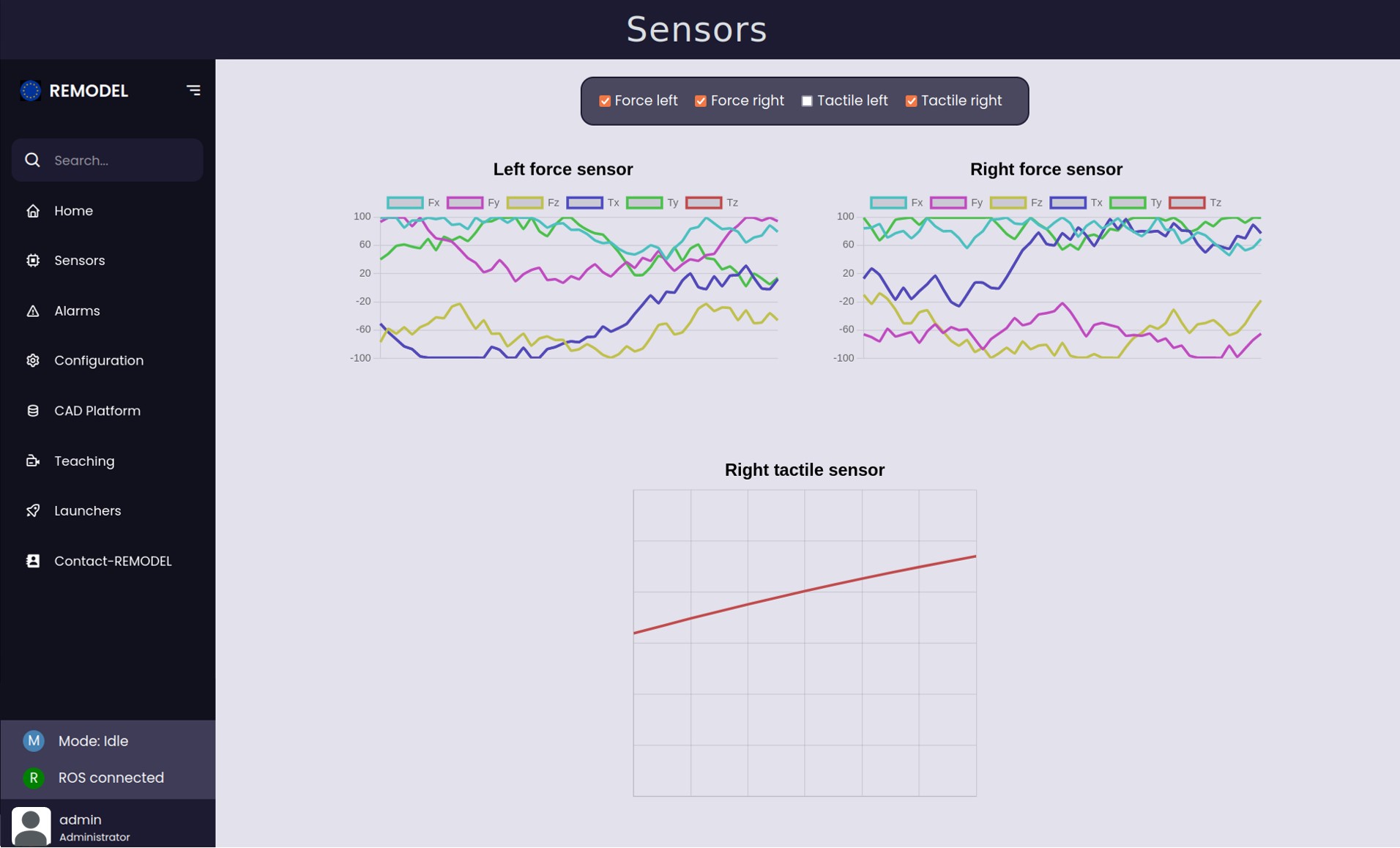}
    \caption{UI Sensor data visualization.}
    \label{fig:sensors}
\end{figure*}

Two ROS services were developed to achieve the functionality described in this subsection, one for launching a list of launch files, and the other for stopping them. These services are called when clicking the ``Launch" or ``Stop" buttons of a module, sending them the list of launch files associated with it. Upon receiving this list, the services check if these files have already been launched and, if applicable, use the roslaunch library to launch or stop them. This approach prioritizes the UI reconfigurability, allowing to define all the modules displayed on the page with a single list, which contains the name of the modules, their nodes and launch files, and the list of user roles that have permission to launch them.

\subsection{Sensor data visualization}

Sensors are essential components in the majority of robotic systems, enabling them to perceive the environment and adjust the robot's motion accordingly. Graphically visualizing the data captured by these sensors can be highly beneficial for users in various scenarios. For example, it can aid in debugging or calibrating the system, identifying errors, or monitoring ongoing processes. The Sensor data visualization feature provides the UI with this functionality, plotting the data of various sensors in multiple graphs, as can be seen in Fig.~\ref{fig:sensors}. These graphs were created utilizing the Chart.js JavaScript library.

The design of this feature has emphasized its scalability and reconfiguration. Thus, the user can modify the sensor graphs to be displayed by editing the elements of a single list. Each of these elements contains information about a sensor graph, such as its title, type (scatter graph or time-evolution graph), axes limits, labels, and the name of the ROS topic that publishes its data. To ensure the sensor's data compatibility, standardized messages are used for the topics publishing their data. Thus, if it is a scatter graph, the message must contain two lists with the values for the X and the Y axis respectively. An example of this is the ``Right tactile sensor" graph in the bottom of Fig.~\ref{fig:sensors}. On the other hand, if it is a time-evolution graph, the message must include two lists, one containing the value of each represented variable, and the other containing their names. An example of this are the force sensor graphs in the top of Fig.~\ref{fig:sensors}, which plot the value of six variables (the forces and torques in the three axes) over time.

\subsection{Robot manual motion control}

This feature allows to manually operate the robotic system from the UI. It provides three options for controlling the robot, achieved by the interaction of the UI with a dedicated ROS node. The first option allows users to command the robot to a pre-defined configuration. As can be seen in Fig.~\ref{fig:manual} (left), the user can select the target robot group and its desired configuration from two dropdown menus. These menus are populated automatically upon mounting the component in the UI by querying a dedicated ROS service running on the manual control node. This service is robot-agnostic, as it requests this information from the RobotCommander MoveIt class, which facilitates the reconfiguration of the UI when a different robot is employed.

\setlength{\textfloatsep}{8pt}
\begin{figure*}[!t]
    \centering
    \includegraphics[width=.98\textwidth]{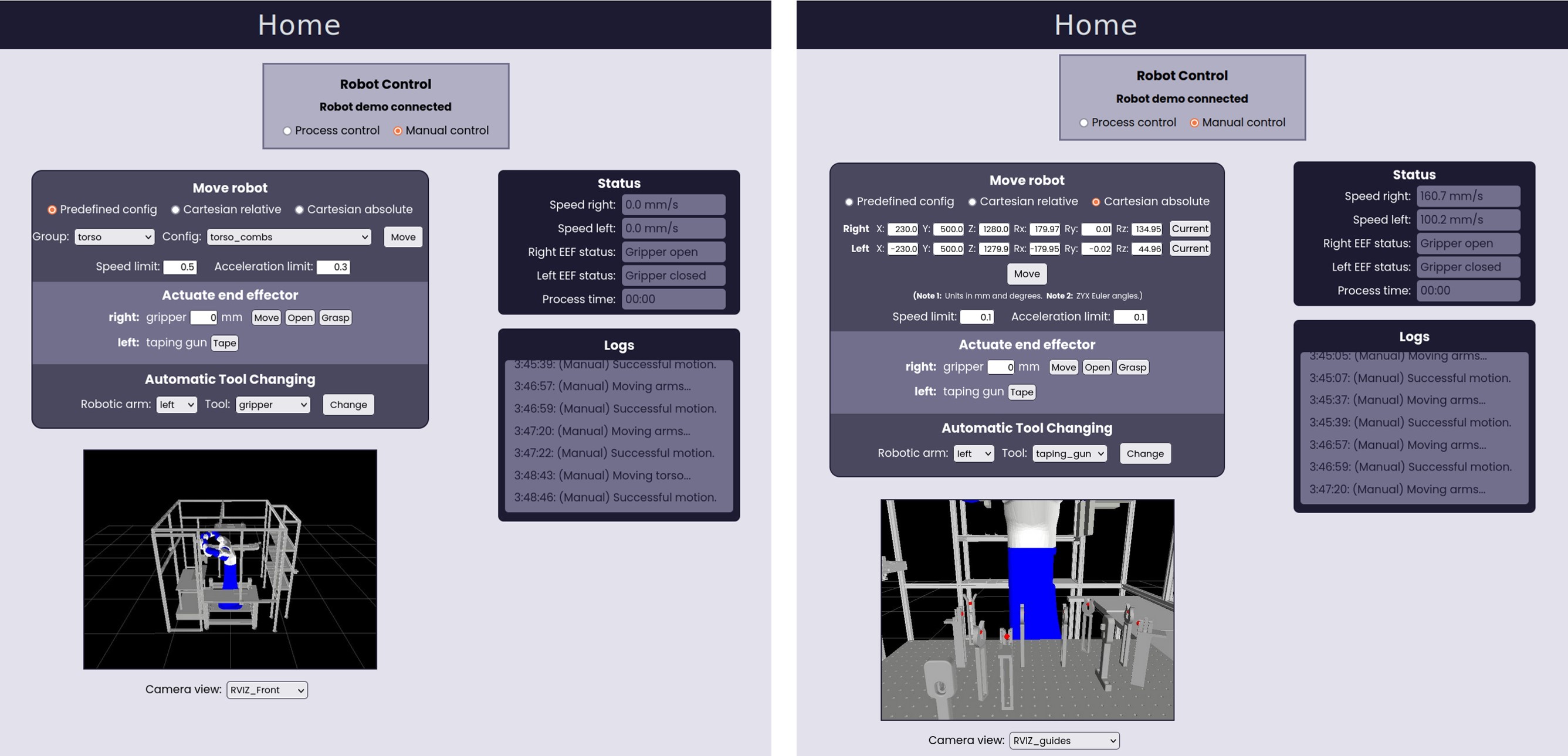}
    \caption{Manual control of the robot from the UI, moving to predefined robot configurations (left) or executing cartesian absolute movements (right).}
    \label{fig:manual}
\end{figure*}

On the other hand, the second and third options enable the user to execute relative and absolute movements of the robotic arms within the Cartesian space, respectively. Fig.~\ref{fig:manual} (right) shows the interface for the third option, where the user can specify the target absolute pose for each arm. The displayed position and orientation of the robotic arms in this figure are retrieved through a dedicated ROS service call. Furthermore, for all three options, users can specify the speed and acceleration limits. Finally, when the “Move” button is clicked, another ROS service is called, which utilizes MoveIt functions to move the robot to the specified target pose or configuration.

Beyond robot movement control, this feature offers functionalities for actuating the robot's end effectors and managing tool changes. As illustrated in Fig.~\ref{fig:manual}, the UI provides a user-friendly interface for interacting with various tools. This particular example showcases two distinct tool types: a parallel gripper and a taping gun. However, this could be easily adapted to different end effectors by simply modifying the topic that is published when clicking its button, so it initiates the intended ROS action. 

\subsection{Robot automatic control and monitoring}

\setlength{\textfloatsep}{8pt}
\begin{figure}[!t]
    \centering
    \includegraphics[width=.95\columnwidth]{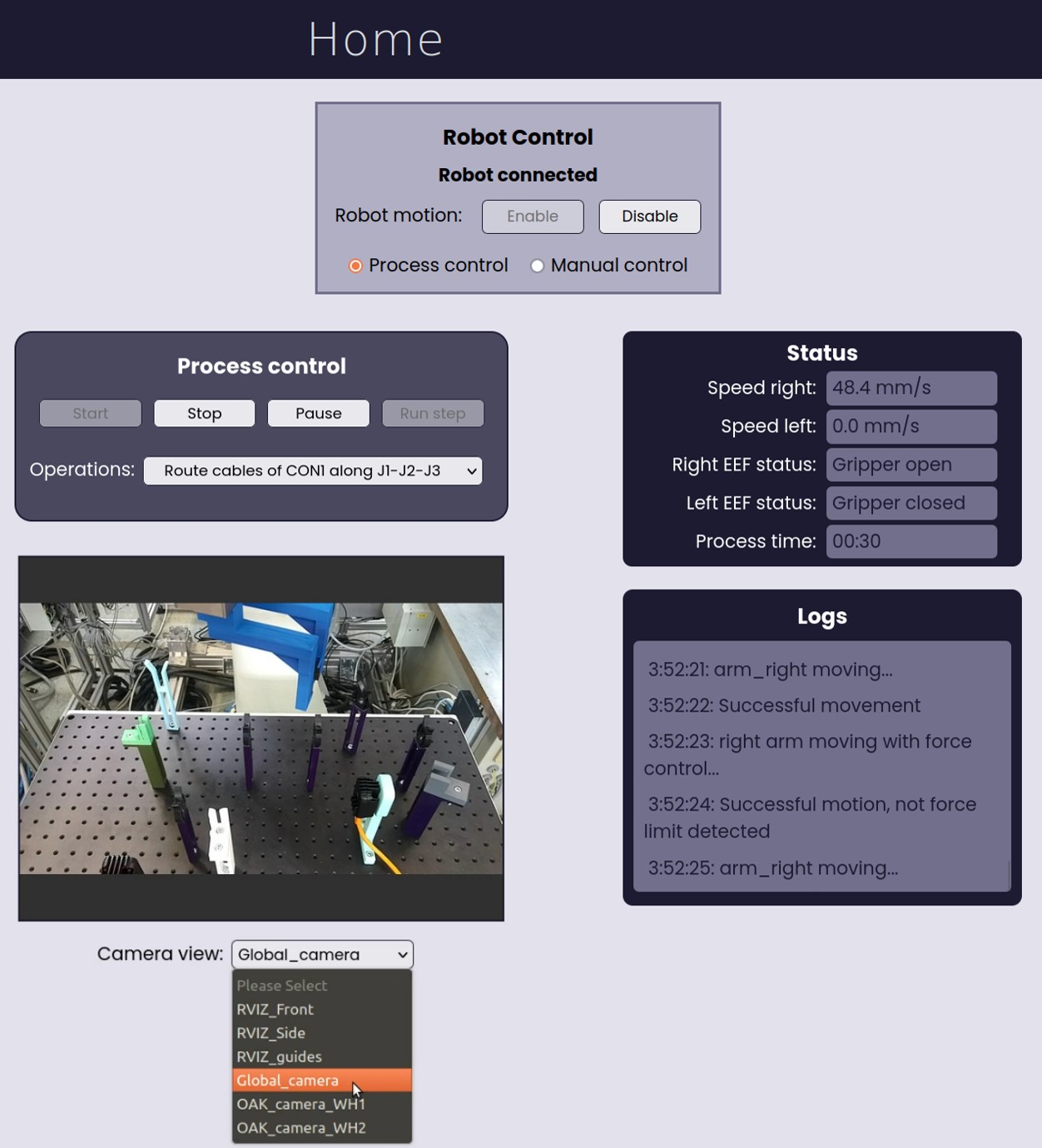}
    \caption{Robot automatic control and monitoring and workbench video stream visualization through the UI}
    \label{fig:auto}
\end{figure}

This feature allows users to control and monitor the robotic system during the automatic or semi-automatic execution of a process. Designed to be versatile, it is not tailored to any specific process but instead provides a generic interface suitable for commanding a variety of processes. As can be seen in Fig.~\ref{fig:auto}, this interface accomplishes two objectives: controlling the robotic system (left side of the page content area) and monitoring the process (on the right). The control section includes four buttons and a dropdown menu, which shows the sequence of operations that compose the process. The first three buttons (i.e., `Start', `Stop', and `Pause/Resume') can be used for commanding the robot in automatic mode, displaying the current operation in execution in the dropdown menu; while the fourth button (i.e., `Run step') can be used for semi-automatic control, enabling the user to just execute the operation selected in the menu. As for the monitoring section, it contains two feedback panels, one for visualizing the status of the system, which shows the values of various variables in real-time, such as the robot speed, or the total process execution time; and the other one for spawning informative messages, such as the success or failure of the robot's movements.

Regarding the back-end functionality of this feature, it is provided by two ROS nodes. The first node functions as a ROS action server, responsible for generating and executing the robot's planned trajectories. On the other hand, the second node acts as a ROS action client, which either calls the action server with specific goals or cancels ongoing actions. Furthermore, this second node is the one responsible for interacting with the UI through different topics, which are published when pressing the four aforementioned buttons. Thus, even if the server node varies for each specific robot application, this will not affect its communication with the UI, as the client node acts as an intermediary between them. Regarding the UI monitoring section, both these nodes can publish messages into any of the feedback panels' topics to update or add relevant information.

\subsection{Video stream visualization}

A potential drawback of using a GUI for interacting with a robotic system is that it can divert users' attention away from the robot's operation, as it is not possible to monitor both simultaneously. To address this limitation, this feature offers a solution by allowing users to visualize video streams from different cameras directly within the UI. This functionality applies to both real cameras positioned within the workstation and virtual cameras integrated into the RVIZ environment as CameraPubs \cite{cameraPub2020Feb}. To transmit their video streams to the UI, each camera must publish its captured image frames to a designated topic at a specified frequency. Conversely, the UI must subscribe to all these topics to receive the stream information from all the cameras. Finally, the video stream to be displayed will be selected by the user from a dropdown menu. Examples of different camera streams can be found in Fig.~\ref{fig:manual} (virtual camera global and workbench views) and Fig.~\ref{fig:auto} (real camera workbench view).

\subsection{System and process configuration}

\setlength{\textfloatsep}{8pt}
\begin{figure}[!t]
    \centering
    \includegraphics[width=.99\columnwidth]{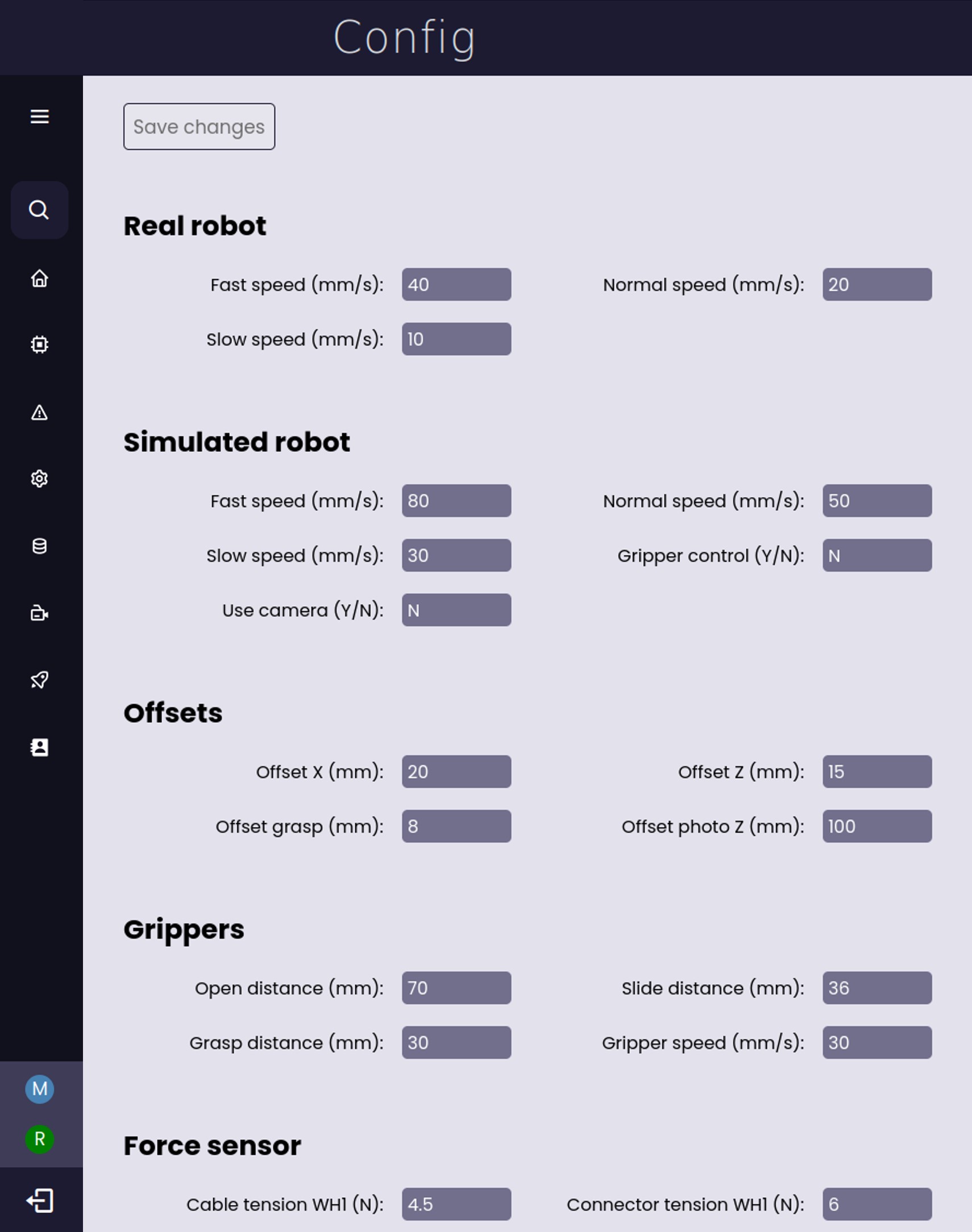}
    \caption{Robotic system and process configuration through the UI}
    \label{fig:config}
\end{figure}

This feature provides an interface to visualize and easily modify system and process-related parameters, such as the desired robot speed levels, force threshold values for force controllers, vision system parameters, or offset values. These parameters are read from and overwritten into a CSV configuration file, which must be queried by any configurable ROS node of the system upon its initialization. This interface allows the organization of the configuration parameters into different categories to enhance its usability, as can be seen in Fig.~\ref{fig:config}. Furthermore, the design of this feature has emphasized its reconfigurability, enabling the easy modification of the editable configuration parameters by updating the fields of a single list.

\setlength{\textfloatsep}{8pt}
\begin{figure}[!t]
    \centering
    \includegraphics[width=.9\columnwidth]{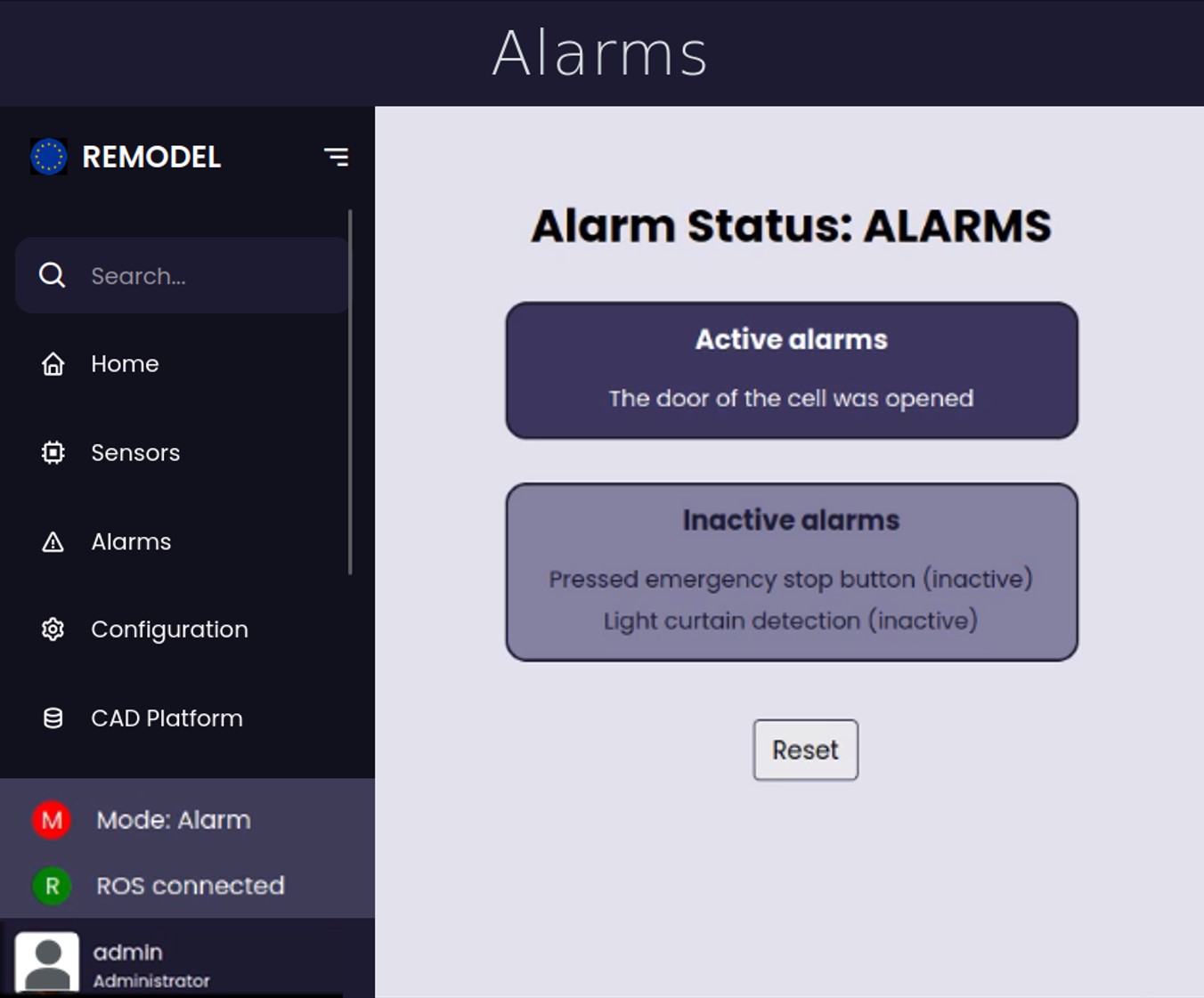}
    \caption{UI alarms visualization}
    \label{fig:alarms}
\end{figure}

\setlength{\textfloatsep}{8pt}
\begin{figure*}[!t]
    \centering
    \includegraphics[width=1.4\columnwidth]{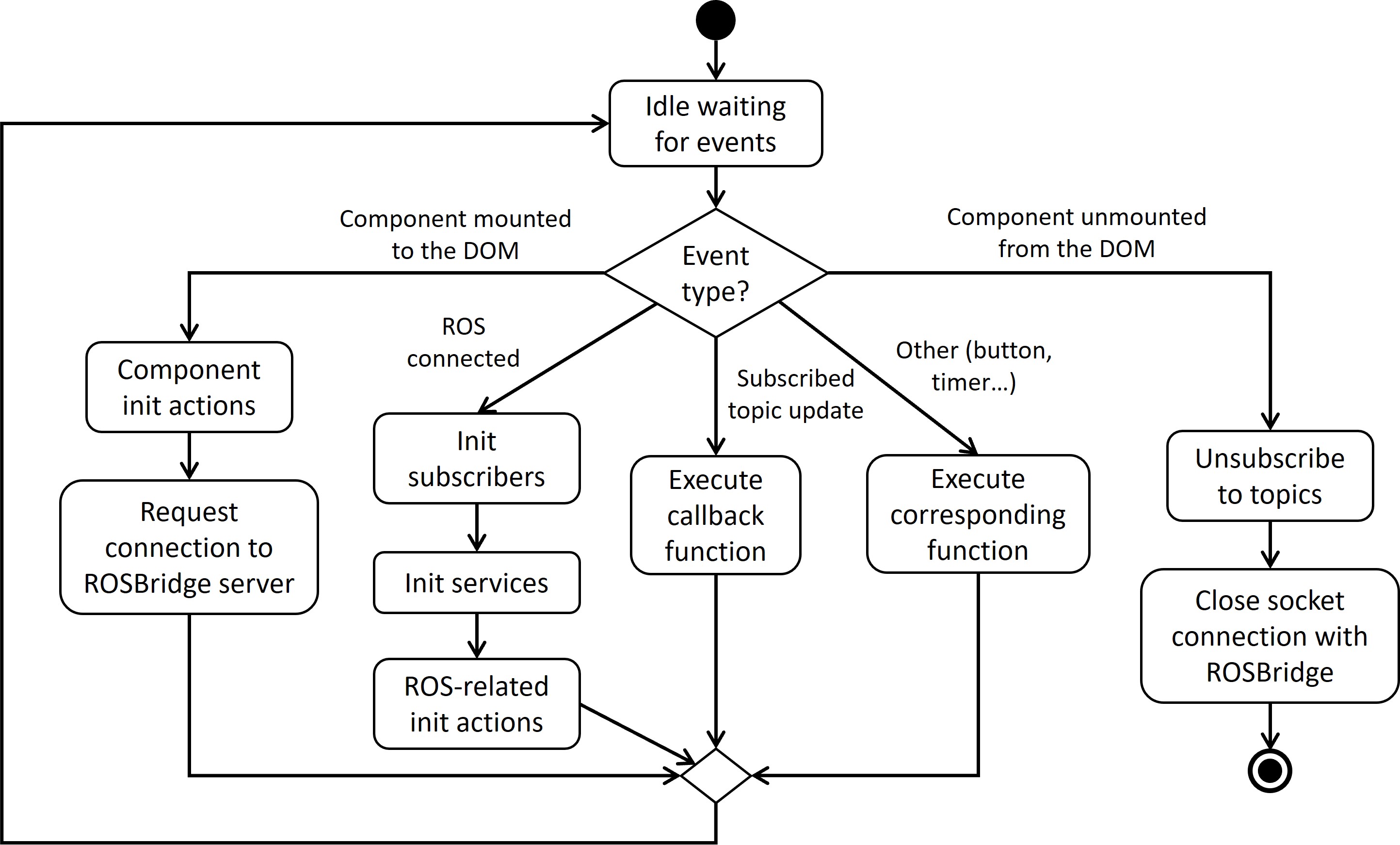}
    \caption{UML activity diagram of the feature components pipeline.}
    \label{fig:pipeline1}
\end{figure*}

\subsection{Trajectory/routine recording and testing}

Besides defining a robot's behavior by using a graphical user interface or by manually coding it in a programming language, there are other more intuitive options. One of these options is to ``teach" the robot a certain routine by recording a sequence of waypoints and actions that the robot has to follow. This is known as \emph{teleoperation}, when the robot is moved to these waypoints without human contact, using a teach pendant, or as \emph{kinesthetic teaching}, when the robot is physically moved by the user using gravity compensation \cite{akgun2012novel}. This feature provides a generic interface for any of these teaching methods. 

This interface contains three panels. The first one is used for recording routines. Here the user can select the robot's motion group and end effector to be recorded, in case of having multiple, and can add poses and end effector actions to the routine. To record a new pose the user must first move the robot to that pose, which can be done by any means (either teleoperation or kinesthetic teaching). This feature utilizes ROS services to record new poses and actions and save the routine. The second panel allows to manage the recorded trajectories, enabling the user to either execute or delete them. These functionalities are provided by ROS services as well. Finally, the third panel includes logs to provide feedback during the process, for instance, ``Pose recorded: [pose coordinates]", ``Grasp recorded", etc.

\subsection{Alarms visualization}

This feature enables bidirectional communication with the node in charge of managing the safety of the system. Thus, as can be seen in Fig.~\ref{fig:alarms}, including it, it is possible to visualize the safety status and the list of alarms of the system, as well as to reset and clear the alarms directly from the UI. This communication is established through ROS topics. Hence, for this to be possible, the node publishing the system's safety information must adhere to a certain message structure, which consists of the list of alarms and their status: active or inactive. The status of an alarm is active when the alarm is still being detected, therefore, it would re-appear immediately in case of resetting; and it is inactive when the alarm is no longer being detected, but the system safety hasn't been reset since it occurred.

\subsection{Databases update}

Most robotic systems use different kinds of databases or information systems to load predefined knowledge or information and use it to configure the robot's trajectories and operations. Therefore, when there are changes in the production orders or the robotic platform layout, this information, or these configuration files, need to be updated. This would require to manually navigate to the configuration files folder and replace them, which in medium/big systems is not intuitive and is prone to errors if the user accidentally deletes the wrong configure file. Hence, this feature provides an interface to overwrite configuration files or entire directories directly from a USB flash drive. This interface enables the user to select one of the connected USB sticks, the file to upload, and the database file to be overwritten. To minimize potential errors, this feature only allows replacing a certain list of files, which must be specified by the user, and it prevents it if the files don't have the same extension. Finally, this feature has been designed to be generic, so it can be employed by any robotic system database by simply updating its path and the list of editable files.  

\subsection{System reconfigurability}
\label{subsec:ROS}

\begin{table*}[t]
\renewcommand{\arraystretch}{1.25}
\caption{Reconfiguration possibilities for each UI feature and their required code modifications}
\label{tab:reconfig}
\begin{tabular}{llp{0.55\textwidth}}
\specialrule{.12em}{.06em}{.06em}
\textbf{Feature}                                                                                    & \textbf{Reconfiguration possibilities}                                                                      & \textbf{Required code modifications}                                                                                                                                                           \\ \specialrule{.12em}{.06em}{.06em}
Multirole security clearance                                                                        & Modify users information                                                                                    & Edit elements of a JSON file (username, encrypted password, role)                                                                                                                              \\ \specialrule{.1em}{.05em}{.05em}
\begin{tabular}[c]{@{}l@{}}Dynamic modules launching \\ and monitoring\end{tabular}                 & \begin{tabular}[c]{@{}l@{}}Add/remove displayed \\ modules\end{tabular}                                     & Edit elements of a list, specifying: name, list of launch files, list of nodes, list of authorized user roles                                                                                  \\ \specialrule{.1em}{.05em}{.05em}
Sensor data visualization                                                                           & Add/remove sensor graphs                                                                                    & Edit elements of a list, specifying: name, id, graph title, graph type, axes limits, labels, sensor's topic                                                                                   \\ \specialrule{.1em}{.05em}{.05em}
\multirow{4}{*}{\begin{tabular}[c]{@{}l@{}}Robot manual motion \\ control\end{tabular}}              
                                                                                                    & \begin{tabular}[c]{@{}l@{}}Add/remove controllable \\ robot arms and end effectors\end{tabular}                             & Edit a list that includes the names of the available tools for each robotic arm                                                                                                                                         \\ \cline{2-3} 
                                                                                                    & \begin{tabular}[c]{@{}l@{}}Modify default motion values\\ (e.g., speed, acceleration...)\end{tabular}       & Edit several variables value                                                                                                                                                                   \\ \cline{2-3} 
                                                                                                    & \begin{tabular}[c]{@{}l@{}}Update the names of topics \\ and services\end{tabular}                          & Edit several variables value                                                                                                                                                                   \\ \cline{2-3} 
                                                                                                    & \begin{tabular}[c]{@{}l@{}}Adapt the functions to the \\ specific robot used\end{tabular}                   & Edit the back-end ROS node                                                                                                                                                                     \\ \specialrule{.1em}{.05em}{.05em}
\multirow{3}{*}{\begin{tabular}[c]{@{}l@{}}Robot automatic motion\\ control\end{tabular}}           & \begin{tabular}[c]{@{}l@{}}Add/remove fields displayed \\ in status panel\end{tabular}                      & Edit elements of a list, specifying: id, displayed name, default value                                                                                                                         \\ \cline{2-3} 
                                                                                                    & \begin{tabular}[c]{@{}l@{}}Update the names of topics \\ and services\end{tabular}                          & Edit several variables value                                                                                                                                                                   \\ \cline{2-3} 
                                                                                                    & \begin{tabular}[c]{@{}l@{}}Adapt the functions to the \\ specific robot used\end{tabular}                   & Edit the back-end ROS node                                                                                                                                                                     \\ \specialrule{.1em}{.05em}{.05em}
Video stream visualization                                                                          & Add/remove video sources                                                                                    & Edit elements of a list, specifying: stream name, image topic                                                                                                                                  \\ \specialrule{.1em}{.05em}{.05em}
\begin{tabular}[c]{@{}l@{}}System and process \\ configuration\end{tabular}                         & \begin{tabular}[c]{@{}l@{}}Add/remove the displayed\\ reconfigurable parameters\end{tabular}                & \begin{tabular}[c]{@{}l@{}}Edit elements of a list, specifying: section, displayed name, param id, default value \\ Edit the list of parameters in the configuration CSV file\end{tabular} \\ \specialrule{.1em}{.05em}{.05em}
\multirow{3}{*}{\begin{tabular}[c]{@{}l@{}}Trajectory/routine recording\\ and testing\end{tabular}} & \begin{tabular}[c]{@{}l@{}}Add/remove displayed \\ motion groups\end{tabular}                               & Edit elements of a list, specifying: motion group name                                                                                                                                         \\ \cline{2-3} 
                                                                                                    & \begin{tabular}[c]{@{}l@{}}Add/remove displayed \\ end effectors\end{tabular}                               & Edit elements of a list, specifying: tool name                                                                                                                                                 \\ \cline{2-3} 
                                                                                                    & \begin{tabular}[c]{@{}l@{}}Adapt the functions to the \\ specific robot used\end{tabular}                   & Edit the back-end ROS node                                                                                                                                                                     \\ \specialrule{.1em}{.05em}{.05em}
Alarms visualization                                                                                & \begin{tabular}[c]{@{}l@{}}Update the name of the \\ topic that publishes the \\ safety status\end{tabular} & Edit a variable value                                                                                                                                                                          \\ \specialrule{.1em}{.05em}{.05em}
\multirow{2}{*}{Databases update}                                                                   & \begin{tabular}[c]{@{}l@{}}Edit the list of files that can\\ be overwritten\end{tabular}                    & Edit elements of a list, specifying: file name                                                                                                                                                 \\ \cline{2-3} 
                                                                                                    & \begin{tabular}[c]{@{}l@{}}Update the path of the \\ database directory\end{tabular}                        & Edit a variable value                                                                                                                                                                          \\ \specialrule{.1em}{.05em}{.05em}
\end{tabular}
\end{table*}
\begin{table*}[t]
\centering
\renewcommand{\arraystretch}{1.2}
\caption{ROS developments required for each UI feature. Topic subscribers (sub) and publishers (pub) are defined from the point of view of the UI.} 
\label{tab:topics}
\begin{tabular}{p{0.195\textwidth}p{0.23\textwidth}p{0.375\textwidth}}
\specialrule{.12em}{.06em}{.06em}
\textbf{Feature}                                                                                     & \textbf{Topics}                                                                                          & \textbf{Services}                                          \\ \specialrule{.12em}{.06em}{.06em}
\multirow{2}{*}{General}                                                                             & Operation mode (sub)                                                                                    & Request operation mode when loading page                   \\
                                                                                                     & Safety status (sub)                                                                                     &                                                            \\ \hline
\multirow{2}{*}{\begin{tabular}[c]{@{}l@{}}Dynamic modules launching \\ and monitoring\end{tabular}} &                                                                                                          & Launch a list of launch files                              \\
                                                                                                     &                                                                                                          & Stop a list of launch files                                \\ \hline
Sensor data visualization                                                                            & Sensors readings (sub)                                                                                  &                                                            \\ \hline
\multirow{3}{*}{\begin{tabular}[c]{@{}l@{}}Robot manual motion \\ control\end{tabular}}              & Actuate EEFs action goal (pub)                                                                           & Request the pose of the robotic arms                       \\
                                                                                                     & Start tool changing (pub)                                                                                & Request the names of the controllable motion groups        \\
                                                                                                     &                                                                                                          & Move a motion group to a certain target                    \\ \hline
\multirow{5}{*}{\begin{tabular}[c]{@{}l@{}}Robot automatic motion\\ control\end{tabular}}            & Current operation index (sub)                                                                           & Request the list of operations that the robot will execute \\
                                                                                                     & Logs panel messages (sub)                                                                               & Enable real robot motion                                   \\
                                                                                                     & Status panel messages (sub)                                                                             & Disable real robot motion                                  \\
                                                                                                     & Robot status (sub)                                                                                      &                                                            \\
                                                                                                     & \begin{tabular}[c]{@{}l@{}}Robot control commands: start,\\ 
    \hspace{2mm} stop, pause, resume, step (pub)\end{tabular} &                                                            \\ \hline
Video stream visualization                                                                           & \begin{tabular}[c]{@{}l@{}}Image frames of each video\\ \hspace{2mm} stream (sub)\end{tabular}                       &                                                            \\ \hline
\multirow{2}{*}{\begin{tabular}[c]{@{}l@{}}System and process \\ configuration\end{tabular}}         &                                                                                                          & Request current configuration values                       \\
                                                                                                     &                                                                                                          & Send updated configurated values                           \\ \hline
\multirow{3}{*}{\begin{tabular}[c]{@{}l@{}}Trajectory/routine recording\\ and testing\end{tabular}}  & Start routine execution (pub)                                                                            & Request the list of saved trajectories                     \\
                                                                                                     &                                                                                                          & Add a new action to the recording routine or save it       \\
                                                                                                     &                                                                                                          & Delete a saved trajectory                                  \\ \hline
\multirow{4}{*}{Alarms visualization}                                                                & Reset alarms (sub)                                                                                      &                                                            \\
                                                                                                     & Reset alarms from UI (pub)                                                                               &                                                            \\
                                                                                                     & Alarms updates (sub)                                                                                    &                                                            \\
                                                                                                     & Request alarms update (pub)                                                                              &                                                            \\ \hline
\multirow{3}{*}{Databases update}                                                                    &                                                                                                          & Request the list of connected USB drives                   \\
                                                                                                     &                                                                                                          & Request the list of files in a USB drive                   \\
                                                                                                     &                                                                                                          & Overwritte a file of the database                          \\ \hline
\end{tabular}
\end{table*}

As it has been highlighted during the entire Section~\ref{sec:features}, the design of this UI and all its features has emphasized reconfigurability and adaptability to different robotic platforms and use cases. This reconfigurability has been implemented at two levels. On a higher level, the UI allows the easy integration of features according to the application needs. This is possible due to the modularity of the features, which are independent of each other and follow a common pipeline that is depicted in Fig.~\ref{fig:pipeline1}. This way, it is very simple to add or remove a feature, as long as it adheres to this structure. On the other hand, on a lower level, every feature has been designed to be highly adaptable and reconfigurable, so they can be easily customized and adapted to different applications. Table~\ref{tab:reconfig} compiles all the manual reconfiguration possibilities for each feature and shows the code modifications required to implement each of them. As can be seen, with minimal code modifications, the user has a lot of freedom to tailor the features. This has been achieved due to the high reconfigurability of the features' methods, allowing the user, in most cases, to entirely customize them by simply updating the value of a few variables.

Finally, it is worth mentioning that the UI is just the front-end component of the system, however, in most cases,  back-end developments are also necessary to achieve the desired functionality of a feature. Table~\ref{tab:topics} shows all the information exchanged between the UI and ROS through topics and services for each feature. Therefore, to integrate a ROS system with the UI, it would be necessary to develop the counterpart communication functions (i.e., topics and services) on the ROS side. Some of these are independent of the application, for instance, the functions that provide the dynamic modules launching and monitoring functionality; however, in most cases, this requires the development of tailored ROS nodes, such as for the robot automatic motion control. Hence, to facilitate the integration, this ROS back-end template repository\footnote{ \href{https://github.com/pablomalvido/ROS_UI_backend}{https://github.com/pablomalvido/ROS\_UI\_backend}} can be used as a reference.
\section{Implementation}
\label{sec:UC}

\setlength{\textfloatsep}{8pt}
\begin{figure*}[!t]
    \centering
    \includegraphics[width=.9\textwidth]{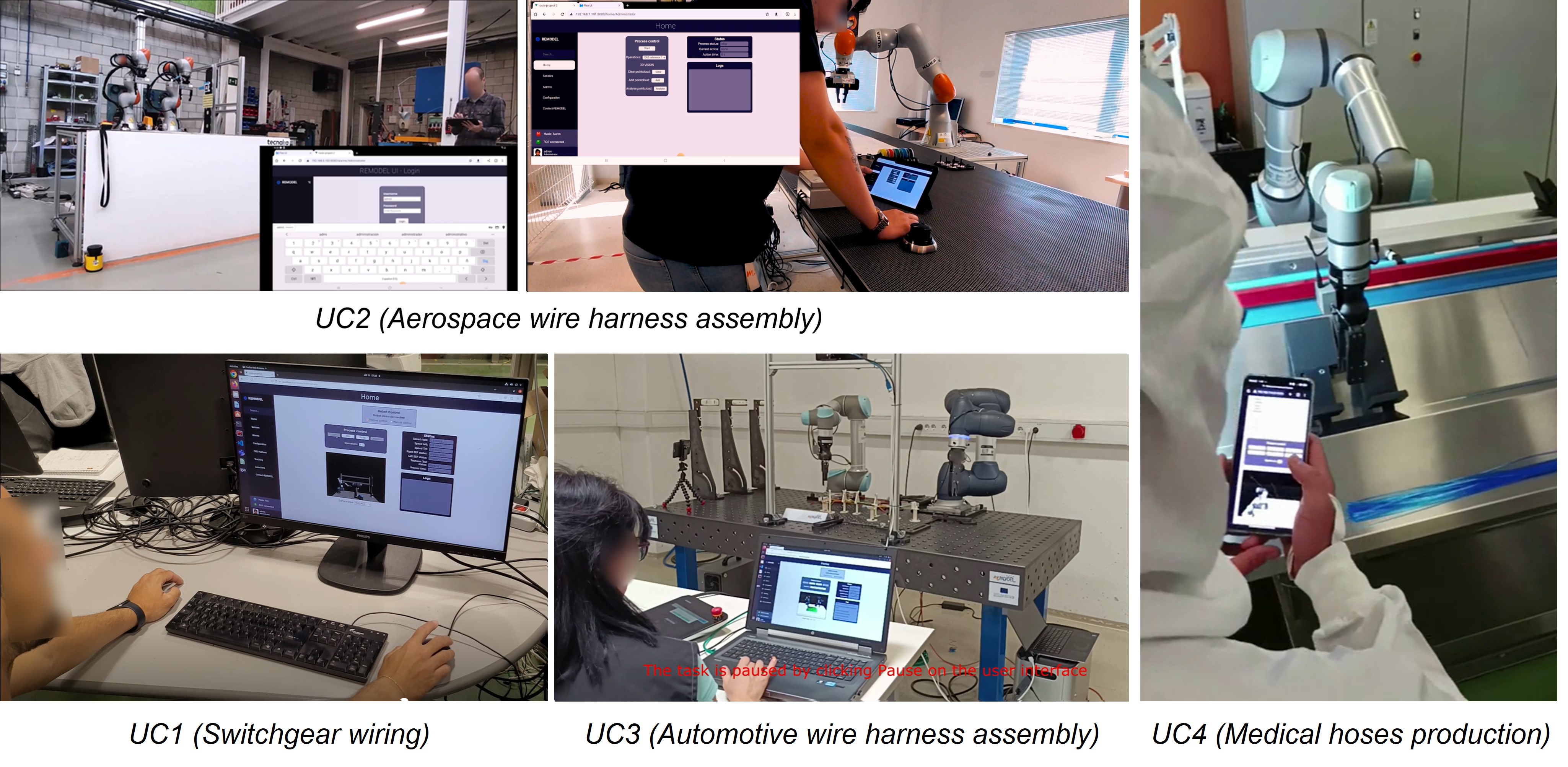}
    \caption{User Interface implementation in four different industrial use cases.}
    \label{fig:remodel}
\end{figure*}

\begin{table}[t]
\renewcommand{\arraystretch}{1.2}
\caption{Set of features incorporated in the UI for its implementation in different use cases} 
\label{tab:features}
\centering
\begin{tabular}{l|cccc}
\hline
\textbf{Feature}                         & \textbf{UC1} & \textbf{UC2} & \textbf{UC3} & \textbf{UC4} \\ \hline
Multirole security clearance             & X            & X            & X            & X            \\
\begin{tabular}[c]{@{}l@{}}Dynamic modules launching and \\ \hspace{0.3cm}monitoring\end{tabular} & X            &              & X            & X            \\
Sensor data visualization                & X            &              & X            &              \\
Robot manual motion control              &              & X            & X            & X            \\
Robot automatic motion control           & X            & X            & X            & X            \\
Video stream visualization               & X            & X            & X            & X            \\
System and process configuration         & X            & X            & X            & X            \\
\begin{tabular}[c]{@{}l@{}}Trajectory/routine recording and \\ \hspace{0.3cm}testing\end{tabular}&              &              &              & X            \\
Alarms visualization                     & X            & X            & X            & X            \\
Databases update                         & X            &              & X            &              \\ \hline
\end{tabular}
\end{table}

As mentioned in the previous sections, one of the main advantages of the proposed UI for ROS systems is its capability to easily adapt to new scenarios, robotic platforms, and applications. To demonstrate this ability, the UI has been implemented and tested in four different robotic use cases (UCs), which are part of the European project REMODEL \cite{Cordis2019Oct}. This project aims to research and develop robotic technologies to handle deformable linear objects (DLOs), such as cables, hoses, or ropes, with robots. Thus, the testing use cases encompass a variety of industrial processes dedicated to the production or assembly of different DLO products: switchgear wiring (UC1), aerospace wire harness assembly (UC2), automotive wire harness assembly (UC3), and medical hoses production (UC4). All these use cases employed different robotic platforms, composed of different number of robots from different vendors, and utilized different devices for visualizing the UI, as can be seen in Fig.~\ref{fig:remodel}. In particular, UC1 utilized two UR5 collaborative robots, and the UI was displayed on a computer; UC2 employed two collaborative KUKA LBR iiwa robots, and a tablet device for the UI; UC3 had two collaborative robots, a UR5 and a Doosan M0609, and a computer for the UI; and UC4 utilized a single UR5 collaborative robot and the UI was displayed on a smartphone. Furthermore, the UI was also used for controlling a Motoman dual-arm industrial robot as can be seen in the figures of Section~\ref{sec:features}.

Due to the different requirements of each use case, they integrated different sets of features into the common UI structure, which have been summarized in Table~\ref{tab:features}. As can be seen, all the features presented in Section~\ref{sec:features} were tested in at least one real use case. The reconfiguration and customization of the features was straightforward in all cases and no problems were reported. Regarding usability, the feedback received was very positive, allowing a seamless interaction with the systems during the development, implementation, and testing phases of the use case demonstrators.
\section{Conclusions}
\label{sec:conclusions}

This paper presents a versatile and reconfigurable UI for controlling, monitoring, and configuring complex ROS-based robotic systems. This UI has a modular design, which facilitates its customization for various systems. This allows to selectively add different features to a generic platform based on the requirements of the application. To ensure seamless integration, added features must adhere to a specific structure. As an example, a set of ten highly relevant features for most robotic applications has been developed. These features have been designed for easy reconfiguration, enabling full customization with minimal changes. The UI's adaptability has been demonstrated through implementation in four industrial use cases involving diverse processes and robotic systems. The results were very positive, significantly improving system usability and presenting no issues during integration.

The main objective of this development is to help the robotic community with a tool that facilitates robotic research and improves the usability of ROS-based systems. Therefore, the repository of the UI, as well as the repository of a reference ROS system integrated with the UI, have been made publicly available. In the future, we will maintain these repositories, offer support for their integration, and extend the current functionalities of the UI by developing new useful features.

\bibliographystyle{IEEEtran}
\bibliography{refs}

\begin{thebibliography}{10}
\providecommand{\url}[1]{#1}
\csname url@samestyle\endcsname
\providecommand{\newblock}{\relax}
\providecommand{\bibinfo}[2]{#2}
\providecommand{\BIBentrySTDinterwordspacing}{\spaceskip=0pt\relax}
\providecommand{\BIBentryALTinterwordstretchfactor}{4}
\providecommand{\BIBentryALTinterwordspacing}{\spaceskip=\fontdimen2\font plus
\BIBentryALTinterwordstretchfactor\fontdimen3\font minus
  \fontdimen4\font\relax}
\providecommand{\BIBforeignlanguage}[2]{{%
\expandafter\ifx\csname l@#1\endcsname\relax
\typeout{** WARNING: IEEEtran.bst: No hyphenation pattern has been}%
\typeout{** loaded for the language `#1'. Using the pattern for}%
\typeout{** the default language instead.}%
\else
\language=\csname l@#1\endcsname
\fi
#2}}
\providecommand{\BIBdecl}{\relax}
\BIBdecl

\bibitem{Wong2022Jun}
C.-C. Wong, C.-Y. Tsai, R.-J. Chen, S.-Y. Chien, Y.-H. Yang, S.-W. Wong, and
  C.-A. Yeh, ``{Generic Development of Bin Pick-and-Place System Based on Robot
  Operating System},'' \emph{IEEE Access}, vol.~10, pp. 65\,257--65\,270, Jun.
  2022.

\bibitem{Nam2020Jun}
C.~Nam, S.~Lee, J.~Lee, S.~H. Cheong, D.~H. Kim, C.~Kim, I.~Kim, and S.-K.
  Park, ``{A Software Architecture for Service Robots Manipulating Objects in
  Human Environments},'' \emph{IEEE Access}, vol.~8, pp. 117\,900--117\,920,
  Jun. 2020.

\bibitem{MalvidoFresnillo2023Oct}
P.~Malvido~Fresnillo, S.~Vasudevan, W.~M. Mohammed, J.~L. Martinez~Lastra, and
  J.~A. Perez~Garcia, ``{Extending the motion planning
  framework{\ifmmode---\else\textemdash\fi}MoveIt with advanced manipulation
  functions for industrial applications},'' \emph{Rob. Comput. Integr. Manuf.},
  vol.~83, p. 102559, Oct. 2023.

\bibitem{FresnilloICPS}
P.~M. Fresnillo, S.~Vasudevan, and W.~M. Mohammed, ``An approach for the
  bimanual manipulation of a deformable linear object using a dual-arm
  industrial robot: cable routing use case,'' in \emph{2022 IEEE 5th
  International Conference on Industrial Cyber-Physical Systems (ICPS)}.\hskip
  1em plus 0.5em minus 0.4em\relax IEEE, 2022, pp. 1--8.

\bibitem{Ajaykumar2021Oct}
G.~Ajaykumar, M.~Steele, and C.-M. Huang, ``{A Survey on End-User Robot
  Programming},'' \emph{ACM Comput. Surv.}, vol.~54, no.~8, pp. 1--36, Oct.
  2021.

\bibitem{Oulasvirta2020Feb}
A.~Oulasvirta, N.~R. Dayama, M.~Shiripour, M.~John, and A.~Karrenbauer,
  ``{Combinatorial Optimization of Graphical User Interface Designs},''
  \emph{Proc. IEEE}, vol. 108, no.~3, pp. 434--464, Feb. 2020.

\bibitem{Coleman2014Apr}
D.~Coleman, I.~Sucan, S.~Chitta, and N.~Correll, ``{Reducing the Barrier to
  Entry of Complex Robotic Software: a MoveIt! Case Study},'' \emph{arXiv},
  Apr. 2014.

\bibitem{Carroll1998}
J.~M. Carroll, ``The evolution of human-computer interaction,'' \emph{Annual
  Review of Psychology}, vol.~48, pp. 501--522, 2001.

\bibitem{Sutherland1998Jul}
I.~E. Sutherland, ``{Sketchpad{\ifmmode---\else\textemdash\fi}a man-machine
  graphical communication system},'' in \emph{{AFIPS '63 (Spring): Proceedings
  of the May 21-23, 1963, spring joint computer conference}}.\hskip 1em plus
  0.5em minus 0.4em\relax New York, NY, USA: Association for Computing
  Machinery, Jul. 1998, vol.~1, pp. 391--408.

\bibitem{Johnson1989Oct}
J.~Johnson, T.~L. Roberts, W.~Verplank, D.~C. Smith, C.~H. Irby, M.~Beard, and
  K.~Mackey, ``{The Xerox Star: a retrospective},'' \emph{Computer}, vol.~22,
  no.~9, pp. 11--26, Sep. 1989.

\bibitem{Lampson1988Jan}
B.~Lampson, ``A history of personal workstations,'' \emph{ed. A. Goldberg,
  Addison-Wesley}, pp. 291--344, 1988.

\bibitem{Myers1998Mar}
B.~A. Myers, ``{A brief history of human-computer interaction technology},''
  \emph{interactions}, vol.~5, no.~2, pp. 44--54, Mar. 1998.

\bibitem{Jansen1998Apr}
B.~J. Jansen, ``{The graphical user interface},'' \emph{SIGCHI Bull.}, vol.~30,
  no.~2, pp. 22--26, Apr. 1998.

\bibitem{Martinez2011Mar}
W.~L. Martinez, ``{Graphical user interfaces},'' \emph{WIREs Comput. Stat.},
  vol.~3, no.~2, pp. 119--133, Mar. 2011.

\bibitem{Seffah2004Oct}
A.~Seffah, P.~Forbrig, and H.~Javahery, ``{Multi-devices
  {\textquotedblleft}Multiple{\textquotedblright} user interfaces: development
  models and research opportunities},'' \emph{Journal of Systems and Software},
  vol.~73, no.~2, pp. 287--300, Oct. 2004.

\bibitem{Majrashi2015ACU}
K.~Majrashi and M.~Hamilton, ``{A Cross-Platform Usability Measurement
  Model},'' \emph{Lecture Notes on Software Engineering}, vol.~3, pp. 132--144,
  May 2015.

\bibitem{Grosskurth2006}
A.~Grosskurth and M.~W. Godfrey, ``Architecture and evolution of the modern web
  browser,'' \emph{University of Waterloo in Canada}, vol.~24, 2006.

\bibitem{Berners-Lee1995Nov}
T.~Berners-Lee and D.~Connolly, ``Hypertext markup language-2.0,'' Tech. Rep.,
  1995.

\bibitem{Bos2006}
\BIBentryALTinterwordspacing
I.~H. Bert~Bos, Tantek~Çelik and H.~W. Lie, ``{Cascading Style Sheets, level 2
  revision 1},'' Apr. 2006, [Online; accessed 19. Apr. 2024]. [Online].
  Available: \url{https://www.w3.org/TR/2006/WD-CSS21-20060411}
\BIBentrySTDinterwordspacing

\bibitem{Iqbal2017May}
J.~Iqbal, Z.~H. Khan, and A.~Khalid, ``{Prospects of robotics in food
  industry},'' \emph{Food Sci. Technol.}, vol.~37, pp. 159--165, May 2017.

\bibitem{Kasina2017Oct}
H.~Kasina, M.~V. A.~R. Bahubalendruni, and R.~Botcha, ``Robots in medicine:
  past, present and future,'' \emph{International Journal of Manufacturing,
  Materials, and Mechanical Engineering (IJMMME)}, vol.~7, no.~4, pp. 44--64,
  Oct. 2017.

\bibitem{Bartos2021Jan}
M.~Barto{\ifmmode\check{s}\else\v{s}\fi}, V.~Bulej,
  M.~Bohu{\ifmmode\check{s}\else\v{s}\fi}{\ifmmode\acute{\imath}\else\'{\i}\fi}k,
  J.~Stan{\ifmmode\check{c}\else\v{c}\fi}ek, V.~Ivanov, and P.~Macek, ``{An
  overview of robot applications in automotive industry},'' \emph{Transp. Res.
  Procedia}, vol.~55, pp. 837--844, Jan. 2021.

\bibitem{Cubero2020Jul}
S.~Cubero, E.~Marco-Noales, N.~Aleixos,
  S.~Barb{\ifmmode\acute{e}\else\'{e}\fi}, and J.~Blasco, ``{RobHortic: A Field
  Robot to Detect Pests and Diseases in Horticultural Crops by Proximal
  Sensing},'' \emph{Agriculture}, vol.~10, no.~7, p. 276, Jul. 2020.

\bibitem{Mohebbi2020Sep}
A.~Mohebbi, ``{Human-Robot Interaction in Rehabilitation and Assistance: a
  Review},'' \emph{Curr. Robot. Rep.}, vol.~1, no.~3, pp. 131--144, Sep. 2020.

\bibitem{Ben-Ari2017Oct}
M.~Ben-Ari and F.~Mondada, ``{Robots and Their Applications},'' in
  \emph{{Elements of Robotics}}.\hskip 1em plus 0.5em minus 0.4em\relax Cham,
  Switzerland: Springer, Oct. 2017, pp. 1--20.

\bibitem{Latikka2021Nov}
R.~Latikka, N.~Savela, A.~Koivula, and A.~Oksanen, ``{Attitudes Toward Robots
  as Equipment and Coworkers and the Impact of Robot Autonomy Level},''
  \emph{Int. J. Social Rob.}, vol.~13, no.~7, pp. 1747--1759, Nov. 2021.

\bibitem{Henneman1999Jan}
R.~L. Henneman, ``Design for usability: Process, skills, and tools,''
  \emph{Information Knowledge Systems Management}, vol.~1, no.~2, pp. 133--144,
  1999.

\bibitem{mikhalevich2017developing}
S.~Mikhalevich, N.~Krinitsyn, F.~Manenti, V.~Kurochkin, S.~Baydali
  \emph{et~al.}, ``Developing of kuka youbot software for education process,''
  \emph{Chemical Engineering Transactions}, vol.~57, pp. 1573--1578, 2017.

\bibitem{quigley2009ros}
M.~Quigley, K.~Conley, B.~Gerkey, J.~Faust, T.~Foote, J.~Leibs, R.~Wheeler,
  A.~Y. Ng \emph{et~al.}, ``Ros: an open-source robot operating system,'' in
  \emph{ICRA workshop on open source software}, vol.~3, no. 3.2.\hskip 1em plus
  0.5em minus 0.4em\relax Kobe, Japan, 2009, p.~5.

\bibitem{Rajapaksha}
D.~D. Rajapaksha, M.~N.~M. Nuhuman, S.~D. Gunawardhana, A.~Sivalingam, M.~N.~M.
  Hassan, S.~Rajapaksha, and C.~Jayawardena, ``{Web Based User-Friendly
  Graphical Interface to Control Robots with ROS Environment},'' in \emph{{2021
  6th International Conference on Information Technology Research
  (ICITR)}}.\hskip 1em plus 0.5em minus 0.4em\relax IEEE, 2021, pp. 01--03.

\bibitem{Barros}
I.~R.~C. Barros, L.~F. Costa, and T.~P. Nascimento, ``{TurtleUI: A Generic
  Graphical User Interface for Robot Control},'' in \emph{{2019 Latin American
  Robotics Symposium (LARS), 2019 Brazilian Symposium on Robotics (SBR) and
  2019 Workshop on Robotics in Education (WRE)}}.\hskip 1em plus 0.5em minus
  0.4em\relax IEEE, 2019, pp. 23--25.

\bibitem{Velamala}
S.~S. Velamala, D.~Patil, and X.~Ming, ``{Development of ROS-based GUI for
  control of an autonomous surface vehicle},'' in \emph{{2017 IEEE
  International Conference on Robotics and Biomimetics (ROBIO)}}.\hskip 1em
  plus 0.5em minus 0.4em\relax IEEE, pp. 05--08.

\bibitem{Mohamed2021}
Y.~Mohamed and S.~Lemaignan, ``Ros for human-robot interaction,'' in \emph{2021
  IEEE/RSJ international conference on intelligent robots and systems
  (IROS)}.\hskip 1em plus 0.5em minus 0.4em\relax IEEE, 2021, pp. 3020--3027.

\bibitem{Bandara}
H.~M. Y. L.~W. Bandara, D.~S. Wijesekera, H.~M. T. D.~B. Herath, D.~L.
  Kodagoda, and S.~Rajapaksha, ``{Methodology for Coping with Uncertain
  Information Contained in Natural Language Instructions in a Robotic
  System},'' in \emph{{2020 2nd International Conference on Advancements in
  Computing (ICAC)}}.\hskip 1em plus 0.5em minus 0.4em\relax IEEE, pp. 10--11.

\bibitem{rajapaksha2019}
S.~Rajapaksha, V.~Illankoon, N.~D. Halloluwa, M.~Satharana, and D.~Umayanganie,
  ``{Responsive Drone Autopilot System for Uncertain Natural Language
  Commands},'' in \emph{{2019 International Conference on Advancements in
  Computing (ICAC)}}.\hskip 1em plus 0.5em minus 0.4em\relax IEEE, pp. 05--07.

\bibitem{Costa2016}
L.~F. Costa and L.~M. Gon{\c{c}}alves, ``Roboserv: A ros based approach towards
  providing heterogeneous robots as a service,'' in \emph{2016 XIII Latin
  American Robotics Symposium and IV Brazilian Robotics Symposium
  (LARS/SBR)}.\hskip 1em plus 0.5em minus 0.4em\relax IEEE, 2016, pp. 169--174.

\bibitem{Hoogervorst2017May}
R.~Hoogervorst, C.~Trouwborst, A.~Kamphuis, and M.~Fumagalli, ``Viki—more
  than a gui for ros,'' \emph{Robot Operating System (ROS) The Complete
  Reference (Volume 2)}, pp. 633--655, 2017.

\bibitem{Casan}
G.~A. Casa{\ifmmode\tilde{n}\else\~{n}\fi}, E.~Cervera, A.~A. Moughlbay,
  J.~Alemany, and P.~Martinet, ``{ROS-based online robot programming for remote
  education and training},'' in \emph{{2015 IEEE International Conference on
  Robotics and Automation (ICRA)}}.\hskip 1em plus 0.5em minus 0.4em\relax
  IEEE, pp. 26--30.

\bibitem{Crick2016Aug}
C.~Crick, G.~Jay, S.~Osentoski, B.~Pitzer, and O.~C. Jenkins, ``{Rosbridge: ROS
  for Non-ROS Users},'' in \emph{{Robotics Research}}.\hskip 1em plus 0.5em
  minus 0.4em\relax Cham, Switzerland: Springer, Aug. 2016, pp. 493--504.

\bibitem{roslibjs}
\BIBentryALTinterwordspacing
``{roslibjs - ROS Wiki},'' Apr. 2024, [Online; accessed 11. Apr. 2024].
  [Online]. Available: \url{http://wiki.ros.org/roslibjs}
\BIBentrySTDinterwordspacing

\bibitem{akahon2024Apr}
\BIBentryALTinterwordspacing
akahon, ``{vue-sidebar-menu-akahon},'' Apr. 2024, [Online; accessed 11. Apr.
  2024]. [Online]. Available:
  \url{https://github.com/akahon/vue-sidebar-menu-akahon}
\BIBentrySTDinterwordspacing

\bibitem{cameraPub2020Feb}
\BIBentryALTinterwordspacing
``{Stream Rviz Visualizations as Images},'' Feb. 2020, [Online; accessed 23.
  Apr. 2024]. [Online]. Available:
  \url{https://roboticsknowledgebase.com/wiki/tools/stream-rviz}
\BIBentrySTDinterwordspacing

\bibitem{akgun2012novel}
B.~Akgun, K.~Subramanian, and A.~L. Thomaz, ``Novel interaction strategies for
  learning from teleoperation,'' in \emph{2012 AAAI Fall Symposium Series},
  2012.

\bibitem{Cordis2019Oct}
\BIBentryALTinterwordspacing
{\relax Cordis}.~{\relax europa}.~e. Cordis, ``{Robotic tEchnologies for the
  Manipulation of cOmplex DeformablE Linear objects},'' \emph{CORDIS {$\vert$}
  European Commission}, Oct. 2019. [Online]. Available:
  \url{https://cordis.europa.eu/project/id/870133}
\BIBentrySTDinterwordspacing

\end{thebibliography}

\end{document}